\documentclass[preprint]{elsarticle}
\usepackage{amsthm}
\usepackage{amssymb}
\usepackage{graphicx}
\usepackage{color,url}
\usepackage{amsmath,amsfonts}
\usepackage{stackrel}
\usepackage{enumitem}
\usepackage{xspace}

\newtheorem{definition}{Definition}
\newtheorem{example}{\textsc{\textbf{Example}}}
\newtheorem{remark}{\textsc{\textbf{Remark}}}
\newtheorem{proposition}{\textsc{\textbf{Proposition}}}
\newtheorem{theorem}{\textsc{\textbf{Theorem}}}
\newtheorem{corollary}{\textsc{\textbf{Corollary}}}
\newtheorem{lemma}{\textsc{\textbf{Lemma}}}
\newtheorem{algorithm}{Algorithm}

\newcommand{\DRa}{\ensuremath{\mathit{DR}_1}}
\newcommand{\DRb}{\ensuremath{\mathit{DR}_2}}
\newcommand{\lang}{\ensuremath{\mathcal{L}}}
\newcommand{\defleftarrow}{{\raise1.5pt\hbox{\tiny\defleft\ \ }}}
\newcommand{\defleft}{\mbox{---\hspace{-2.5pt} \raise-0.07pt\hbox{$<$}}}

\newcommand{\Argu}{\ensuremath{\mathcal{A}}}
\newcommand{\ourRelPrefEst}{\mbox{$\succ$}}
\newcommand{\ourRelPrefInd}{\mbox{$\sim$}}

\newcommand{\intd}[2]{\ensuremath{\langle \{#1\},#2\rangle}}
\newcommand{\inti}[1]{\ensuremath{\mathit{int}(#1)}}
\newcommand{\args}[1]{\ensuremath{\mathit{args}(#1)}}
\newcommand{\ArbD}[1]{$\mathbb{T}_{{\scriptscriptstyle
			\mathcal{K}_{\mathcal{A}}}}($#1$)$}
\newcommand{\EA}[1]{\ensuremath{\Sigma_{#1}}}

\newcommand{\domEA}{\mbox{$\mathfrak{str}_{{\scriptscriptstyle \epistcompabs}}$}}
\newcommand{\epistcompabs}{\mbox{$\mathcal{K}$}}
\newcommand{\derrEAact}{\mbox{$=$\hspace{-5pt}$\gg$}}
\newcommand{\conflictEA}{{\scriptsize \mbox{\raise2.1pt\hbox{$\smile$}\hspace{-3mm}\raise-1.5pt\hbox{$\frown$}}}}

\newcommand{\DLP}{\mbox{DeLP}}
\newcommand{\SADR}{\mbox{$\Gamma$}}
\newcommand{\alternatives}[1]{\Omega_{#1}}

\newcommand{\desirule}[4]{\ensuremath{#1\stackrel{\ \scriptscriptstyle{#2}}{\Leftarrow} #3, \ #4}}
\newcommand{\epistcomp}{$\mathcal{K}$}

\newcommand{\MPAB}{\mbox{$C^*(B,\succsim)$}}
\newcommand{\ourRelPref}{\mbox{$\succsim$}}

\newcommand{\eg}{\emph{e.g.},\xspace}
\newcommand{\etal}{\emph{et al.}\xspace}
\newcommand{\CS}{$(\mathcal{B},C(\cdot))$}
\newcommand{\BSet}{$\mathcal{B}$}
\newcommand{\CDot}{$C(\,\cdot\,)$}
\newcommand{\CB}{$C(B)$}
\newcommand{\ourB}{$B$}
\newcommand{\ie}{\textnormal{\emph{i.e.}, }}

\newcommand{\facto}[1]{\mbox{$ #1$}}

\newcommand{\SAP}{\mbox{$X$}}

\newcommand{\Dcomp}{\ensuremath{\Gamma}}
\newcommand{\run}{{r}}
\newcommand{\lits}{\ensuremath{\mathit{dlits}}}
\newcommand{\DistLit}{\ensuremath{\mathcal{C}}}
\newcommand{\OrderC}{\ensuremath{>_{\DistLit}}}
\newcommand{\ADeF}{$\langle X, \DistLit, \OrderC, \epistcompabs, \Dcomp \rangle$}
\newcommand{\ADeFrun}{$\langle X_\run, \DistLit_\run, \ensuremath{>_{\DistLit_r}}, \epistcompabs_\run, \Dcomp_\run \rangle$}

\begin{document}
\begin{frontmatter}

\title{An approach to Decision Making based on Dynamic Argumentation Systems}

\maketitle
\noindent
\author[rvt]{Edgardo Ferretti}
\ead{ferretti@unsl.edu.ar}
\author[focal]{Luciano H. Tamargo}
\ead{lt@cs.uns.edu.ar}
\author[focal]{Alejandro J. Garc\'ia}
\ead{ajg@cs.uns.edu.ar}
\author[rvt]{Marcelo L. Errecalde}
\ead{merreca@unsl.edu.ar}
\author[focal]{Guillermo R. Simari}
\ead{grs@cs.uns.edu.ar}

\address[rvt]{Department of Informatics, Universidad Nacional de San Luis,  Argentina}
\address[focal]{Institute for Computer Science and Engineering (UNS--CONICET)\\
Department of Computer Science and Engineering, Universidad Nacional del Sur,   Argentina\\
\texttt{http://dx.doi.org/10.1016/j.artint.2016.10.004}}

\begin{abstract}
In this paper we introduce a formalism for single-agent decision making that is based on Dynamic Argumentation Frameworks.
The formalism can be used to justify a choice, which is based on the current situation the agent is involved.
Taking advantage of the inference mechanism of the argumentation formalism, it is possible to consider preference relations, and conflicts among the available alternatives for that reasoning.
With this formalization, given a particular set of evidence, the justified  conclusions supported by warranted arguments will be used by the agent's \textit{decision rules} to determine which alternatives will be selected.
We also present an algorithm that implements a choice function based on our formalization.
Finally, we complete our presentation by introducing formal results that relate the proposed framework with approaches of classical decision theory.
\end{abstract}
\end{frontmatter}

\section{Introduction}
\label{sec:intro}

Argumentation systems are based on the construction and evaluation of interacting
arguments that are intended to support, explain, or attack statements which can be decisions,
opinions, etc. Argumentation has been applied to different domains~\cite{Sim:09}, such as non-monotonic reasoning, handling inconsistency in knowledge bases, and modeling different kinds of dialogues, in particular persuasion and negotiation. An argumentation-based approach to negotiation has the advantage that in addition to the exchange of offerings, also provides reasons to support these offerings. 
In this way, adopting this kind of approach to decision problems has the benefit that
besides choosing a proper alternative, the decision maker could also ponder the underlying reasons supporting the decision in a more understandable manner. That is, giving explanations and justifications of the choices in terms of arguments is more informative and more open to discussion and criticism than referring to a formula for a utility function~\cite{Neu:53}.
Therefore, the idea of articulating decisions based on arguments became relevant to different approaches to decision making, such as decision under
uncertainty~\cite{Amg:06b},
multi-criteria decision~\cite{Oue:07},
rule-based decisions~\cite{Kak:03},
and case-based decisions~\cite{Bru:03}.

Following this trend, we propose an 
approach to single-agent decision making based on Dynamic Argumentation Frameworks.
Dynamic Argumentation Frameworks (or DAF for short) were introduced in~\cite{RotsteinMGS10} and provide a formalization for abstract argumentation systems where the current set of \textit{evidence} dynamically activates arguments that belong to a \textit{working set} of arguments.
The main objective of DAFs is to extend Argumentation Frameworks~\cite{Dung:95} to provide the ability of handling dynamics; for achieving that, at a given moment, the set of available \textit{evidence}  determines which arguments are \textit{active} and can be used to make inferences to obtain justified conclusions.

In our proposal of an Abstract Decision Framework, a DAF is used for representing preference relations and the conflicts among the available alternatives. Four other components complete the formalism:
a set of mutually exclusive \textit{alternatives} which are available to the agent;
a set of \emph{distinguished literals} representing different binary preference relations for comparing the alternatives;
a \emph{strict total order} over the set of distinguished literals to represent the priority among the preference criteria provided to the agent; and
a set of \emph{decision rules} that implement the agent's decision making policy. 
Taking advantage of the argumentation formalism, preference relations and conflicts among the available alternatives can be considered for that reasoning.
With this formalization, given a particular set of evidence, the justified  conclusions supported by warranted arguments will be used by the agent's \textit{decision rules} to determine which alternatives will be selected.
We will also introduce an algorithm that implements a choice function based on our formalization.
We will complete the presentation introducing formal results that relate the choice behavior 
of our proposed framework 
to Classical Decision Theory~\cite{Arrow:59,Mas:95,Ritcher:71}.

In classical approaches to decision making, the objectives of a decision maker are summarized in a \emph{rational preference relation}, or in a \emph{utility function} representing this relation. 
Despite the criticisms received, expected utility theory 
has become `the major paradigm in decision making'~\cite{Sch:82}.
As suggested by Parsons and Fox in~\cite{Par:96}, this may be due to the solid theoretical underpinning that numerical methods have; 
that is why they have stated that when developing decision making models based on argumentation formalisms, a key issue is to formally relate them to classical approaches to decision theory.

A particular feature of our approach is that the formalism is not attached to any particular agent architecture.
Since it is based on a dynamic abstract argumentation framework,  some elements of the formalism can be instantiated with a particular argumentation system.
Another feature of our proposal is that using a DAF allows us to apply our framework to environments where the scenario (\ie the available evidence) can change dynamically.

Next, we include an application example that will serve two purposes: to motivate the main ideas of our proposal and as a running example to be used in the rest of the paper.

\begin{example}\label{ex.running}
	In this application the domain consists of a mobile robotic agent that performs a cleaning task.
	In this environment, the agent has to decide which box has to be carried next to a defined area called $store$.
	Boxes can be of different sizes, can be spread over the environment, and each box represents an alternative to be chosen.
	
	\begin{figure}[h]
		\centering
		\includegraphics[width=12cm]{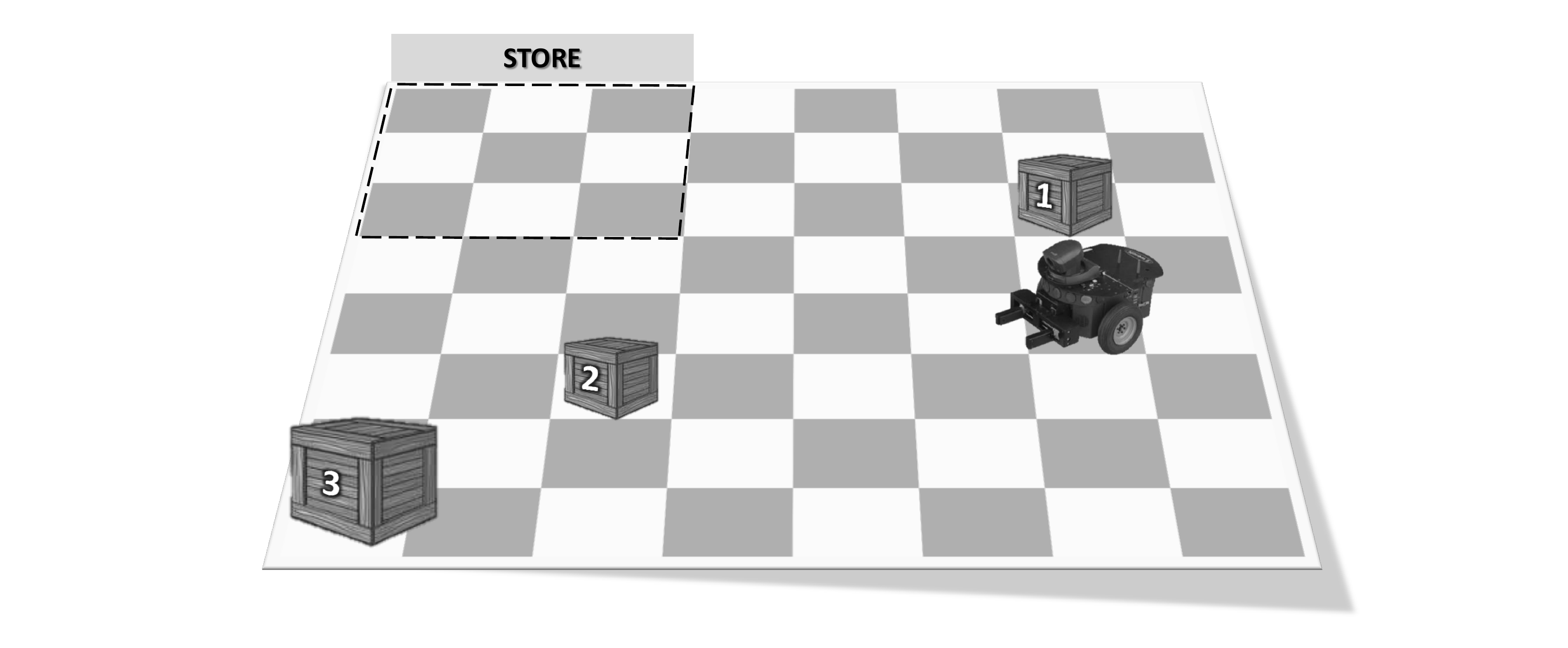}
		\caption{Running example: mobile robotic agent has to decide which box to move next.}
		\label{fig.outline}
	\end{figure}
	
	The agent has a set $\mathbf{P}=\{p_1,p_2,p_3\}$ with three possible preference criteria to compare alternatives:
	$p_1$ for representing that the robot will prefer to choose a small box over a bigger one, $p_2$ for representing that a box nearer to the store will be preferred, and $p_3$ for representing that the robot prefers a box that is near to itself.
	A strict total order over these criteria is also considered: the agent first will prefer boxes nearer to it, \ie  \emph{nearer boxes}, then \emph{boxes nearer to the store}, and finally the \emph{smaller ones}.
	Figure~\ref{fig.outline} shows a particular scenario of our application domain where the robot has to decide among three alternatives: $ box_1$, $box_2$, and $box_3$.
	Observe that  $ box_1$ and $box_2$ have the same size and both are smaller than $box_3$; $ box_1$ is closer to the robot than the other two;
	$box_2$ is closer to the robot than $box_3$;
	$ box_2$ is closer to the store than the other two;
	and that $box_1$ is closer to the store than $box_3$.
\end{example}

Consider a scenario as the one introduced in Example~\ref{ex.running} where the robot is faced with three alternatives.
Following our proposal, as will be explained below, this application domain will be represented with a dynamic argumentation framework that will provide: a working set of arguments, a way of representing the conflict among these arguments, and a preference relation for deciding between conflictive arguments.
Then, the available evidence of a particular scenario will be used to identify the arguments that are active, and these arguments will be used to obtain justified conclusions with respect to this evidence.
In the situation of Example~\ref{ex.running}, as will be shown, our proposed framework will select $box_1$.

As it was mentioned before, the use of argumentative reasoning for decision making has been studied in other approaches such as~\cite{Amg:09,Atk:06,JETAI2014,Par:96}), and the  subject will be thoroughly analyzed in Section~\ref{sec:related} where the related work is discussed.
In particular,~\cite{JETAI2014} proposes a framework to represent the agent's preferences and its knowledge using Possibilistic Defeasible Logic Programming~\cite{Alsi:08}; there, warranted information is used in decision rules that implement the agent's decision-making policy.
In contrast to \cite{JETAI2014}, we develop a more general proposal since our formalization is based on an abstract dynamic argumentation framework that allows the creation of different instantiations.
Like~\cite{JETAI2014}, in this article we propose to use decision rules and an algorithm for computing the acceptable alternatives; however, both elements were reformulated with respect to~\cite{JETAI2014}.

The article is organized as follows.
In Section~\ref{sec:theory} we will introduce the basic concepts of decision-making from the point of view of the standard theory of individual rationality, 
and the development of the contribution of this work spans from Section~\ref{sec:framework} to Section~\ref{sec:formal}.
In Section~\ref{sec:framework} we will present the main results of our approach which consists of an abstract framework for decision making based on decision rules and dynamic argumentation;
we will also introduce in that section the formalization of the epistemic component of our abstract decision framework.
We will present the algorithm for selecting acceptable alternatives in Section~\ref{sec:accepting}.
In Section~\ref{sec:formal} we will  lay down a formal comparison of the choice behavior of the proposed framework with respect to Classical Decision Theory; furthermore, we will discuss how our proposal is related to other significant
decision-making approaches in Section~\ref{sec:related}.
Finally,  we will offer the conclusions and consider future work in Section~\ref{sec:conclusion}.

\section{Preliminaries}
\label{sec:theory}

We will introduce here a brief overview of the theory of individual decision making as presented in~\cite{Mas:95}, where two related approaches to model the agent's decision are considered.
Later on, in Section~\ref{sec:formal}, the choice behavior of the argumentation-based decision framework proposed in this paper, will be formally related with these approaches to model the agent's decisions.

The starting point for any individual decision problem in classical approaches to decision making is the characterization of a set of possible (mutually exclusive) \emph{alternatives} from which the decision maker (an agent in our case) must choose.
Following the notation commonly used in the literature this set of alternatives will be denoted by $X$.
For instance, consider the example we have already introduced (see Figure~\ref{fig.outline}) where the robot has three alternatives to select: $box_1$, $box_2$ or $box_3$; in that case, $X$ becomes $\{box_1, box_2, box_3\}$.

In classical decision making approaches it is usually assumed the agent's choice behavior is modeled with a binary \emph{preference relation} $\succsim$, where given $\{x, y\} \subseteq X$, $x \succsim y$  means that
``$x$ is at least as good as $y$''.
As usual, from $\succsim$ it is possible to derive two other important relations:
\begin{itemize} \itemsep 0pt
	\item The \emph{strict preference} relation $\succ$, defined	as $x \succ y \Leftrightarrow x  \succsim y \mbox{ but not } y \succsim x$ which is read as ``$x$ is preferred to $y$'';
	\item The  \emph{indifference} relation $\thicksim$, defined as $x \thicksim y \Leftrightarrow x \succsim y
	\mbox{ and } y \succsim x$ and which is read as ``$x$ is indifferent to  $y$''.
\end{itemize}

It is customary to require the preference relation $\succsim$ to be \emph{rational} (see definition~\ref{def:rational}) and this becomes a necessary condition when $\succsim$ is represented by a \emph{utility} function.
The hypothesis of rationality is embodied in two basic assumptions concerning the preference relation $\succsim$ as defined next.
\begin{definition}[Rational preference relation]\label{def:rational} A preference relation
	$\succsim$ is rational if it verifies the following two properties:\hspace{1pt}\footnote{In Order Theory, a binary relation satisfying these properties is called a \emph{total pre-order}.}
	\begin{enumerate} \itemsep 0pt
		\item Completeness: for all $x, y \in X$,  $x \succsim y$ or $y \succsim x$ (or both).
		\item Transitivity: for all $x, y, z \in X$, if $x \succsim y$ and $y \succsim z$, then $x \succsim z$.
	\end{enumerate}
\end{definition}

The assumption that $\succsim$ is complete requires that the agent has a well-defined preference between any two possible alternatives.
Also, transitivity implies that it is impossible for the decision maker to be faced with a sequence of pairwise choices in which her preferences appear to cycle.
For instance,  following our running example we can consider that the robot will prefer to select a box that is near to itself, then, in the scenario described in Figure~\ref{fig.outline}: $box_1 \succsim box_2$, $box_1 \succsim box_3$, and $box_2 \succsim box_3$.

Considering the choice behavior of a decision maker with a rational preference relation \ourRelPref\ over \SAP, when facing a non-empty set of alternatives $B \subseteq X$, her preference-maximizing behavior will choose any of the elements in the following set:
\MPAB~$=\{ x \in B~|~x \succsim y$ for each $y \in B \}$.
Following our running example, if we consider $B=\{box_2, box_1\}$ then
\MPAB~$=\{box_1\}$, however, if we consider $B=\{box_2, box_3\}$ then
\MPAB~$=\{box_2\}$.
If $B=\{box_2, box_1, box_3\}$ then \MPAB~$=\{box_1\}$.

It is a well known fact in decision theory community that completeness and transitivity assumptions are usually hard to satisfy in real-world problems (see \eg~\cite{Sim:72,tversky1986rational}), still the \emph{preference-based approach} (PBA) is very relevant from a theoretical point of view.
In fact, this approach is the most traditional way of modeling individual choice behavior.
Nonetheless, the \emph{choice-based approach} introduced next is an interesting proposal which is a more flexible formal model of theory of decision making, since it is based on entirely behavioral foundations rather than being limited to consider individual decision making as an introspection-based process.

The \emph{choice-based approach} (CBA for short)~\cite{Ritcher:71}
takes the choice behavior of the
individual as a primitive object which is represented by means of a \textit{choice structure} $($\BSet,~\CDot$)$ consisting of two elements:
\begin{itemize}\itemsep 0pt
	\item \BSet\ is a set of subsets of $X$.
	Intuitively, each set $B \in$ \BSet\ represents a set of 	alternatives (or \emph{choice experiment}) that can be conceivably posed to the decision maker.
	In this way, if $X=\{x,y,z\}$ and $\mathcal{B}= \{\{x,y\},\{x,y,z\}\}$ we will assume that the sets $\{x,y\}$ and $\{x,y,z\}$ are valid choice experiments to be presented to the decision maker.
	\item \CDot\ is a \emph{choice rule} which basically assigns to each set of alternatives \ourB\ $\in \mathcal{B}$ a non-empty set that represents the alternatives that the decision maker \emph{might} choose when presented
	the alternatives contained in \ourB.
	Note that \CB\ $\subseteq B$ for every \ourB\ $\in	\mathcal{B}$.
	When \CB\ contains a single element, this element represents the \emph{individual's choice} among the alternatives in \ourB.
	The set \CB\ might, however, contain more than one element and in this case they would represent the \emph{acceptable alternatives} in \ourB\ for the agent.
\end{itemize}

\begin{example}\label{ex.CBA}
	Returning to the scenario shown in Figure~\ref{fig.outline}
	where $X$=$\{box_1, box_2,$ $box_3\}$.
	Consider the set $\mathcal{B}=$ $\{\{box_1\},$ $\{box_2\},$ $\{box_3\},$ $\{box_1,box_2\},$ $\{box_1,box_3\},$ \\ $\{box_2,box_3\},$ $\{box_1,box_2,box_3\}\}$.
	Suppose that the choice experiment is $B=\{box_1,box_3\}$.
	If \CB\ = $\{\ box_1\}$ then $box_1$ is the individual's choice.
\end{example}

As in the rationality assumption of the PBA (definition~\ref{def:rational}), in the CBA there is a central
assumption, called the \emph{weak axiom of revealed preference} (or WARP for short)~\cite{Sam:38}.
As it will be explained next, this axiom imposes an element of consistency on choice behavior that is similar to  the
rationality assumptions of the PBA.
The following definition recalls the WARP axiom:
\begin{definition}[Weak axiom of revealed preference]\label{def:weak-ax}A choice structure \CS\ satisfies the weak axiom of revealed preference (WARP) if the following property holds:
	If for some $B \in \mathcal{B}$ with $x,y \in B$ we have $x \in	C(B)$, then for any $B' \in \mathcal{B}$ with $x,y \in B'$ and $y \in C(B')$, we must also have $x \in C(B')$.
\end{definition}

The weak axiom requires that if there is some choice experiment $B \in \mathcal{B}$ such that  $x$ and $y$ are presented as alternatives ($x,y \in B$) and ``$x$ is \emph{revealed at least as good as} $y$'' (\ie $x \in C(B)$) then there does not exist another choice experiment $B^\prime \in \mathcal{B}$  where ``$y$ is \emph{revealed strictly preferred} to $x$'' (\ie $x,y \in B^\prime$,  $y \in C(B^\prime)$ and $x \notin C(B^\prime)$).

Intuitively, the WARP principle reflects the expectation that an individual's observed choices will display a certain amount of coherence.
That is to say, in our example, if given $X=\{box_1, box_2, box_3\}$,
\mbox{$\mathcal{B}=\{\{box_1, box_2\},\{box_1, box_2, box_3\}\}$} and a choice rule $C$, $C(\{box_1, box_2\})=\{box_1\}$, then the axiom says that it cannot be the case that $C(\{box_1, box_2, box_3\})=\{box_2\}$.
In fact, it says more: we must have $C(\{box_1, box_2, box_3\})=\{box_1\}$, or
$C(\{box_1, box_2, box_3\})=\{box_3\}$, or $C(\{box_1, box_2, box_3\})=\{box_1,box_3\}$.

As mentioned above, the PBA and CBA approaches present different perspectives of the theory of individual decision making.
The former considers it as a process of introspection while the latter makes assumptions about objects that are directly observable (choice behavior) rather than things that are not (preferences).
In spite of these differences, under certain conditions these two approaches are related.
Below, we introduce a well known and important result which states that if a decision maker has a rational preference ordering \ourRelPref, when faced with a choice experiment, her choices will necessarily generate a choice structure that satisfies the WARP principle:\\[4pt]
\emph{Suppose that \ourRelPref\ is a rational preference relation then the choice structure generated by
	\ourRelPref, $(\mathcal{B},C^*(\,\cdot\,,\succsim))$ satisfies the weak axiom of revealed preference.}

\section{Argumentation-based Abstract Framework for Decision Making}
\label{sec:framework}

We will now introduce the main contribution of our approach:
an abstract framework for decision making based on decision rules and dynamic argumentation.
Later, in Section~\ref{sec:formal} our proposal will be formally related to Classical Decision Theory.

Our abstract decision framework will integrate five components:
a set of all the \textit{available alternatives} \SAP\ that the decision maker has, a set of \textit{distinguished literals}
that refer to the agent's preferences, a \emph{strict total order} over the set of distinguished literals to represent the priority among the preference criteria provided to the agent,
an \textit{epistemic component} that will be used for representing preference relations and conflicts among the available alternatives, and a \textit{decision component} that effectively implements the agent's decision making policy based
on \emph{decision rules}.
Next, we will define our framework and we will devote the rest of this section to present the epistemic  component, while the decision component will be developed and explained in the following section.
Since this is an abstract framework, the representation language will not be instantiated; however, we assume that this representation language includes constants, predicates, and classical negation ($\neg$).
In our formalization, will refer to this base language as \lang.

\begin{definition}[Abstract Decision Framework]\label{def:abst:df}
	An  \emph{abstract decision framework} for a language \lang, is a tuple \ADeF $_{\lang}$  where:
	\begin{itemize}
		\item $X \subset$ \lang\  is the set of all possible \textit{alternatives};
		\item $\DistLit \subset$ \lang\ is a  \textit{set of distinguished literals};
		\item   \OrderC\ is a strict total order among elements of \DistLit;
		\item $\epistcompabs$, which is referred to as the \textit{epistemic component}, is a dynamic argumentation framework; and
		\item $\Gamma$, called the \textit{decision component}, is a set of decision rules.
	\end{itemize}
\end{definition}

When no confusion could possibly arise we will drop the subindex \lang\ from \ADeF $_{\lang}$ to simplify notation.
The  epistemic component $\epistcompabs$, that will be introduced next, will include evidence and arguments that the decision maker will use for reasoning, and the set $\Gamma$, explained in the next section,  will be used for implementing the agent's decision making policy.

Following our running example we can define the abstract decision framework \ADeFrun, with the set of alternatives $X_\run=\{box_1, box_2, box_3\}$.
The epistemic component $\epistcompabs_\run$ (explained in detail below) will contain evidence obtained from the domain  (\eg  $smaller(box_1,box_3)$, $smaller(box_2,box_3)$) and arguments for and against considering one box better than another one.
These arguments will be considered by the dynamic argumentation framework
to obtain the agent's conclusions (\eg $better(box_1,box_3)$).
This epistemic component provides a knowledge representation tool  that allows for the representation of preferences and conflicts among the agent's available alternatives.


In what follows, we will explain  how to formalize the epistemic component of our abstract decision framework
using a Dynamic Argumentation Framework.
DAFs were introduced in~\cite{RotsteinMGS10} and provide a formalization for abstract argumentation systems
where the current set of \textit{evidence} dynamically activates arguments that belong to a \textit{working set} of arguments.
DAFs  have been  defined as a specialization of Dung's argumentation framework (AF)~\cite{Dung:95}, with the main objective of extending AFs to handle dynamics.
To cope with this, at any given moment, the set of available evidence will determine which arguments are active becoming usable in producing inferences.
In contrast, in Dung's approach the consideration of a changing set of active arguments would involve passing from a framework to another.
To keep our presentation as self-contained as possible, we refer the interest reader to our Appendix where a concise but complete description of DAFs is included.

The remainder of this section will be dedicated to explain how to build a particular DAF that will  formalize the epistemic component for our abstract decision framework that will be responsible of obtaining the agents' conclusions.
This component will consider all the evidence the agent is in possession regarding the current situation of environment which can change dynamically upon perception; different instances of this set of evidence will determine different instances of the DAF.
The epistemic component will also handle the working set of arguments the decision maker will use for reasoning to obtain conclusions; for that reason, preference relations and conflicts among the available alternatives should be considered.
The working set of arguments contains every argument that could be available to be used in the reasoning process.
According to the available set of evidence, there is a subset of this working set that contains the arguments that are active in function of the currently present evidence.

In order to make a decision an agent equiped with our framework may have at its disposal one or more
preference criteria which will be used to compare the alternatives in \SAP; an important requirement is that no preference criterion generates cyclic preferences.
All these preference criteria  will be considered to define the preference function $\mathfrak{pref}$ for the DAF of the epistemic component.

\begin{example}\label{example.preference.criteria}
	Continuing with our running example, we will consider three preference criteria: the robot prefers to choose a small box over a bigger one, the robot prefers to choose a box nearer to the store over a box far to store, and the robot prefers to choose a box that is near to itself.
\end{example}

In our framework, each preference criterion will be associated with a literal in \lang.
These literals  will be called \textit{distinguished} as defined next.

\begin{definition}[Set of distinguished literals]
	\label{def:lit:dist} Let \SAP\ be a set of alternatives.
	The set of distinguished literals  $\DistLit = \{c_1, \ldots, c_n\}$  is a non-empty set of  
	literals from \lang.
	Each literal in $\DistLit$  represents a different binary preference relation for comparing alternatives in \SAP.
\end{definition}

Given a distinguished literal $c \in \mathcal{C}$ and $x,y \in X$ then $c(x,y)$ means that ``$x$ is preferred to $y$ with respect to the preference criterion $c$''.
Consider the comparison criteria mentioned in Example \ref{example.preference.criteria}, in our running example the set of distinguished literals will be $\mathcal{C}_{\run} = \{nearer\_robot,$ $nearer\_store,$ $smaller\}$; thus, considering the scenario depicted in Figure~\ref{fig.outline}, it holds $smaller(box_1,box_3)$.
In our formalization, the literal $same\_att(x,y)$ states that alternatives $x,y \in X$ have the same attribute values with respect to every $c \in \DistLit$;
\eg $same\_att(box_4,box_5)$ means that the two boxes have the same attributes
for each preference criterion of the agent.
Note that in our running example (Figure~\ref{fig.outline}) there are no pair of alternatives with the same attribute values; however, in Example~\ref{example_sameprop} below we will introduce a scenario where there are two boxes with the same attribute values for all the preference criteria.

Since an agent can have more than one preference criterion represented by elements of the set of distinguished literals \DistLit, then in our formalization we consider a strict total order  \OrderC\ among these elements.
If the pair $(c^\prime,c) \in\ \OrderC$ then the criterion represented by the distinguished literal $c^\prime$ is considered better than the one represented by $c$.
\begin{example}\label{ex.dist.lit.and.order}
	Consider the comparison criteria mentioned in Example \ref{example.preference.criteria}.
	Then, for our running example the set of distinguished literals is $\mathcal{C}_{\run} = \{nearer\_robot,$ $nearer\_store,$ $smaller\}$.
	We will use the following total order for representing the agent's preferences among $\mathcal{C}_{\run}$: \\[4pt]
	\hspace*{24pt}$>_{\mathcal{C}_{\run}} = \{(nearer\_robot,$ $nearer\_store),(nearer\_robot,$ $smaller),$\\
	\hspace*{56pt}$(nearer\_store,$ $smaller)\}$.\\[4pt]
	That is, in our running example the agent will have three criteria available, and it will prefer first its \emph{nearer boxes}, then \emph{boxes nearer to the store}, and finally the \emph{smaller ones}.
\end{example}

Next, in Definition~\ref{def:epist:comp:abs}, we will introduce the formalization of the epistemic component \epistcompabs\ of Definition~\ref{def:abst:df} with a particular DAF described as $\langle E, W, \bowtie, \mathfrak{pref} \rangle$. In a DAF (see the Appendix), the set of evidence $E$ may change dynamically upon perception, and different instances of this set  will determine different instances of the DAF; thus, for a given set $E$ there will be a subset of the working set of arguments $W$ that contains the arguments that are active.
In our approach, the set $E$ will contain a snapshot of all the information relative to the current relations among the alternatives of the set \SAP\ with respect to the preference criteria represented in $\mathcal{C}$, \eg $nearer\_robot(box1, box2)$ belongs to the evidence in the scenario depicted in Figure~\ref{fig.outline}.

The working set $W$ will contain arguments for reasoning about when an alternative is better than other.
Note that in a DAF,  an argument $\mathcal{A}$ is a reasoning step for a claim $\alpha$ from a set of premises $\{\beta_1, \ldots, \beta_n\}$ denoted as the pair $\langle \{\beta_1, \ldots, \beta_n\}, \alpha \rangle$.
An argument will be \emph{active} if its premises are satisfied based on the current evidence. Given an evidence set $E$, an argument's premise is satisfied whether it belongs to $E$, or it is the conclusion of an active argument according to $E$.
In this DAF the set $\bowtie$ will contain the conflicts among arguments in $W$.
Given an argument $\mathcal{A} \in W$,
$cl(\mathcal{A})$ denotes the claim of $\mathcal{A}$ and
$\overline{cl({\mathcal{A}})}$ represents
the complement of $cl(\mathcal{A})$ with respect to negation
($\neg$).
Finally, the preference function $\mathfrak{pref}$
will consider all the agents' criteria represented in $\mathcal{C}$,
and, if it is possible, it will return the argument that is based on a better distinguished literal with respect to the order  $>_{\mathcal{C}}$.
Since our epistemic component is defined in an abstract form,
the function $\mathfrak{pref}$ is defined in terms
of  \emph{argumental structures} (denoted with $\Sigma$)  which are built with one or more arguments from $W$ (see Appendix).
In order to compare two argumental structures, distinguished literals will be used.  

\begin{definition}[Epistemic component]
	\label{def:epist:comp:abs} Let $X$ be the set of all the possible candidate alternatives, $\mathcal{C}$ be a set of distinguished literals in $\mathcal{L}$ and $>_{\mathcal{C}}$ be a strict total order over $\mathcal{C}$. An epistemic component \epistcompabs, is a DAF $\langle
	E, W, \bowtie, \mathfrak{pref} \rangle$ where:
	\begin{itemize}\itemsep 6pt
		\item[$\centerdot$] The evidence $E$ is a consistent set of sentences of the form $same\_att(x,y)$ or $c(x,y)$, such that  $x,y \in X$ and $c \in \mathcal{C}$.
		\item[$\centerdot$] The working set $W$ will be such that if $c \in \mathcal{C}$,
		$\{x,y\} \subseteq X\ (x \neq y)$ and $better \not \in \mathcal{C}$ then:
		\begin{equation*}
			\begin{split}
				& \langle \{c(x,y)\}, better(x,y) \rangle \in W \\
				& \langle \{c(x,y)\}, \neg better(y,x) \rangle \in W \\
				& \langle \{c(y,x)\}, better(y,x) \rangle \in W \\
				& \langle \{c(y,x)\}, \neg better(x,y) \rangle \in W\\
				& \langle \{same\_att(x,y)\}, \neg better(x,y) \rangle \in W\\
				& \langle \{same\_att(x,y)\}, \neg better(y,x) \rangle \in W \\
			\end{split}
		\end{equation*}
		\item[$\centerdot$] $\bowtie$\hspace{2pt}$=\{(\mathcal{A}, \mathcal{B})| \{\mathcal{A}, \mathcal{B}\} \subseteq W,
		cl(\mathcal{A})=\overline{cl({\mathcal{B}})}\}$. 
		\item[$\centerdot$]
		Let $\Sigma_1$ and $\Sigma_2$ be two argumental structures in $W$,  then \\[4pt]
		$\mathfrak{pref}(\Sigma_1,\Sigma_2)=
		\left\{
		\begin{array}{ll}
		\Sigma_1 &  if~\forall c \in \lits(\Sigma_2),
		~\exists c^\prime \in \lits(\Sigma_1)\
		\mathrm{st.}~(c^\prime,c) \in >_{\mathcal{C}},\\
		\Sigma_2 &  if~\forall c \in \lits(\Sigma_1),
		~\exists c^\prime \in \lits(\Sigma_2)\
		\mathrm{st.}~(c^\prime,c) \in >_{\mathcal{C}}\\
		\epsilon &  \mathit{otherwise} \end{array}
		\right.\\[4pt]
		$ where $\lits(\Sigma) \subseteq \mathcal{C}$ is the set of distinguished literals that are contained in arguments of an argumental structure $\Sigma$.
	\end{itemize}
\end{definition}
\noindent
In the preceding definition, each argument in $W$ has a set of premises containing only one element:
a distinguished literal $c \in \mathcal{C}$ comparing alternatives $x,y \in X$, such that $c(x,y)$ means that ``$x$ is preferred to $y$''; or a literal $same\_att(x,y)$ stating that alternatives $x,y \in X$ have the same attribute values for each preference criterion provided to the agent.
Given a pair of alternatives $x,y \in X~(x \neq y)$, the conclusion of each argument in $W$ states that: ``$x$ is better than $y$'' ($better(x,y)$) or
``$x$ is not better than $y$'' ($\neg better(x,y)$) or ``$y$ is better than $x$'' ($better(y,x)$) or ``$y$ is not better than $x$'' ($\neg better(y,x)$).

A Dynamic Argumentation Framework imposes that all arguments must be \textit{coherent},  that is, any argument  $\mathcal{A}$ in $W$ must  satisfy:
$\overline{cl(\mathcal{A})} \notin pr(\mathcal{A})$,
$\overline{cl(\mathcal{A})} \notin E$,
$cl(\mathcal{A}) \notin pr(\mathcal{A})$, and
$cl(\mathcal{A}) \notin E$ (see Definition \ref{def:argu:coher:nico} in Appendix).
The following proposition shows that in an epistemic framework built as stated by Definition~\ref{def:epist:comp:abs}, since the literal $better \not \in \mathcal{C}$, all arguments in the working set are coherent.

\begin{proposition}
	\label{prop:argu:cohe:W} Given an epistemic component
	$\epistcompabs = \langle E, W, \bowtie, \mathfrak{pref} \rangle$, all the arguments in $W$ are coherent.
	\begin{proof}
		Straightforward from Definition~\ref{def:epist:comp:abs}.
	\end{proof}
\end{proposition}

As it is explained in detail in the Appendix, when two arguments are in conflict, the function $\mathfrak{pref}$ will determine which one is a defeater of the other.
Then, an argument will be warranted with respect to a DAF if it has no warranted defeater.
A conclusion is justified if it is the conclusion of a warranted argument.

\begin{example}\label{exampleEpistemicComponent}
	Consider the running example where $\SAP_\run=\{box_1,box_2,box_3\}$.
	In Example~\ref{ex.dist.lit.and.order}, we have introduced the set $\mathcal{C}_{\run} = \{nearer\_robot,$ $nearer\_store,$ $smaller \}$ and $>_{\mathcal{C}_{\run}}$ = $\{(nearer\_robot,$ $nearer\_store),$ $(nearer\_robot,$ $smaller),$ $(nearer\_store,$ $smaller)\}$.
	Then, $\epistcompabs_\run=\langle E_\run, W_\run, \bowtie_\run, \mathfrak{pref}_\run \rangle$ is the epistemic component for the scenario depicted in Figure~\ref{fig.outline} (where the robot $r$ is currently involved) with the following set of evidence:
	\[ E_\run=\left\{
	\begin{tabular}{l l}
	$smaller(box_1,box_3),$&$nearer\_store(box_2,box_3),$\\
	$smaller(box_2,box_3),$&$nearer\_robot(box_1,box_2),$\\
	$nearer\_store(box_1,box_3),$&$nearer\_robot(box_1,box_3),$\\
	$nearer\_store(box_2,box_1),$&$nearer\_robot(box_2,box_3)$
	\end{tabular}
	\right\}
	\]

	Given the set of alternatives  $\SAP_\run$ and the distinguished literals $\mathcal{C}_{\run}$, then, following Definition \ref{def:epist:comp:abs}, the working set $W_\run$ has forty-two arguments
	that are depicted with triangles in Figure~\ref{fig:W}. White triangles denote the subset of $W_\run$
	which are active with respect to $E_\run$, and  black triangles
	are the inactive arguments of $W_\run$ with respect to $E_\run$.
	The text above each triangle is the conclusion of the argument, and the text below a triangle is its premise.
	The label inside white triangles will be used for referencing the argument along the paper.
	A solid arrow that connects two active arguments represents that the argument at the beginning of the arrow
	defeats (see Definition~\ref{def:EA:derrota} in the Appendix) the argument at the arrow's end.
	In Figure~\ref{fig:W},  $same\_att$, $smaller$,
	$nearer\_store$, $nearer\_robot$, $better$, $\neg better$,
	$box_1,$ $box_2,$ $box_3$ are abbreviated as: $sa$, $sm$, $ns$, $nr$, $b$,  $\neg b$, $1$, $2$ and $3$ respectively.
	
	\begin{figure}[ht]
		\begin{center}
			\includegraphics[width=115mm]{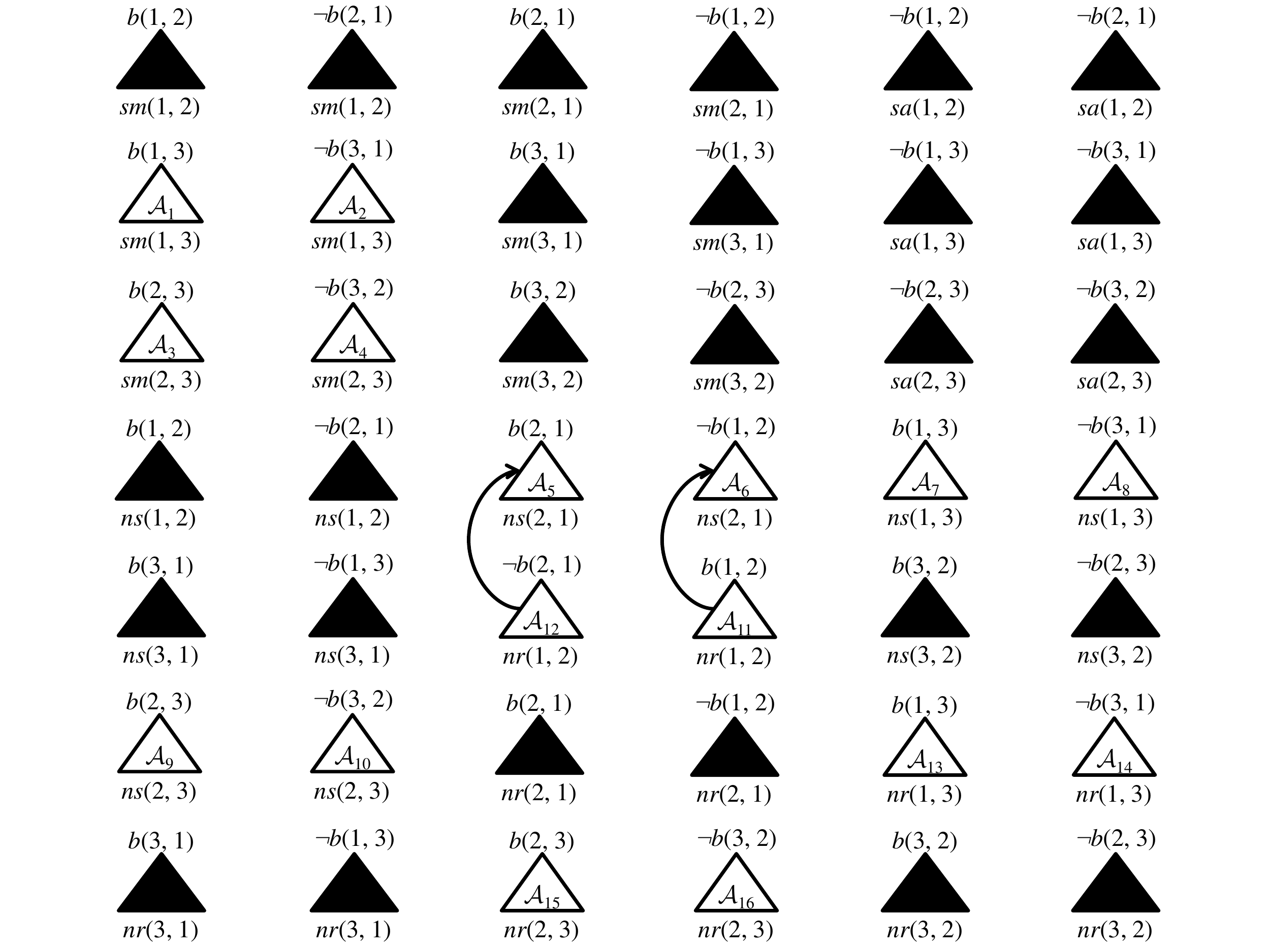}
			\caption{Active and inactive arguments from the working set
				$W_r$ of Example~\ref{exampleEpistemicComponent}.} \label{fig:W}
		\end{center}
	\end{figure}
	
	Given the working set $W_\run$ depicted in Figure~\ref{fig:W} and considering the evidence set $E_\run$ we obtain the following list of active arguments:
	\begin{center}
		\begin{tabular}{lrl}
			$\mathcal{A}_1  $&$=$&$\langle \{smaller(box_1,box_3)\}, better(box_1,box_3)\rangle$,\\
			$\mathcal{A}_3 $&$=$&$\langle \{smaller(box_2,box_3)\}, better(box_2,box_3)\rangle$,\\
			$\mathcal{A}_5 $&$=$&$\langle \{nearer\_store(box_2,box_1)\}, better(box_2,box_1)\rangle$,\\
			$\mathcal{A}_7 $&$=$&$\langle \{nearer\_store(box_1,box_3)\}, better(box_1,box_3)\rangle$,\\
			$\mathcal{A}_9 $&$=$&$\langle \{nearer\_store(box_2,box_3)\}, better(box_2,box_3)\rangle$,\\
			$\mathcal{A}_{11} $&$=$&$\langle \{nearer\_robot(box_1,box_2)\}, better(box_1,box_2)\rangle$,\\
			$\mathcal{A}_{13} $&$=$&$\langle \{nearer\_robot(box_1,box_3)\}, better(box_1,box_3)\rangle$,\\
			$\mathcal{A}_{15} $&$=$&$\langle \{nearer\_robot(box_2,box_3)\}, better(box_2,box_3)\rangle$,\\
			$\mathcal{A}_2 $&$=$&$\langle \{smaller(box_1,box_3)\}, \neg better(box_3,box_1)\rangle$,\\
			$\mathcal{A}_4 $&$=$&$\langle \{smaller(box_2,box_3)\}, \neg better(box_3,box_2)\rangle$,\\
			$\mathcal{A}_6 $&$=$&$\langle \{nearer\_store(box_2,box_1)\}, \neg better(box_1,box_2)\rangle$,\\
			$\mathcal{A}_8 $&$=$&$\langle \{nearer\_store(box_1,box_3)\}, \neg better(box_3,box_1)\rangle$,\\
			$\mathcal{A}_{10} $&$=$&$\langle \{nearer\_store(box_2,box_3)\}, \neg better(box_3,box_2)\rangle$,\\
			$\mathcal{A}_{12} $&$=$&$\langle \{nearer\_robot(box_1,box_2)\}, \neg better(box_2,box_1)\rangle$,\\
			$\mathcal{A}_{14} $&$=$&$\langle \{nearer\_robot(box_1,box_3)\}, \neg better(box_3,box_1)\rangle$,\\
			$\mathcal{A}_{16} $&$=$&$\langle \{nearer\_robot(box_2,box_3)\}, \neg better(box_3,box_2)\rangle$.
		\end{tabular}
	\end{center}
	
	\noindent
	In the list above, those active arguments supporting that an alternative is \textit{better} than other appear first and  those supporting that an alternative is \textit{not better} than other appear later.
	Note that the only conflicts among active arguments are  $ \bowtie_{\run} = \{(\mathcal{A}_6,\mathcal{A}_{11}),
	(\mathcal{A}_{11},\mathcal{A}_{6}), (\mathcal{A}_5,\mathcal{A}_{12}), (\mathcal{A}_{12},$ $\mathcal{A}_{5})\}$.

	Observe that argument $\mathcal{A}_{11}$ supports that $box_1$ is better than $box_2$, based on the premise that $box_1$ is nearer to the robot than $box_2$;
	whereas  $\mathcal{A}_{6}$ supports that $box_1$ is not better than $box_2$ based on the premise that $box_2$ is nearer to the store than $box_1$.
	In the particular case of our running example, all argumental structures have only one argument
	which has one distinguished literal as a premise, hence, for all arguments $\mathcal{A}_i \in W_\run$,
	the set  $\lits(\mathcal{A}_i)$ will be a singleton.
	For instance:  $\lits(\mathcal{A}_{6}) = \{nearer\_store(box_2,box_1)\}$ and  $\lits(\mathcal{A}_{11}) = \{nearer\_robot(box_1,box_2)\}$.
	Since $(nearer\_robot,nearer\_store) \in$ $>_{\mathcal{C}_{\run}}$
	(\ie  the robot  prefers to carry \emph{boxes near to itself}
	than \emph{boxes close to the store})
	then,  $\mathfrak{pref}_\run(\mathcal{A}_{11}, \mathcal{A}_{6}) = \mathcal{A}_{11}$.
	Thus, $\mathcal{A}_{11}$ defeats $\mathcal{A}_{6}$ but $\mathcal{A}_{6}$ does not defeat $\mathcal{A}_{11}$
	(see Definition \ref{def:EA:derrota}).
	Therefore, $\mathcal{A}_{11}$ is warranted in $\epistcompabs_\run$
	and hence, the conclusion $better(box_1,box_2)$ is justified in $\epistcompabs_\run$.
	
	Arguments $\mathcal{A}_{5}$ and $\mathcal{A}_{12}$ are also in conflict, and we have that $\lits(\mathcal{A}_{5}) = \{nearer\_store(box_2, box_1)\}$ and
	$\lits(\mathcal{A}_{12}) = \{nearer\_robot(box_1,box_2)\}.$
	Since $(nearer\_robot, nearer\_store) \in$ $>_{\mathcal{C}_{\run}}$
	then,  $\mathfrak{pref}_\run(\mathcal{A}_{5}, \mathcal{A}_{12}) = \mathcal{A}_{12}$;
	hence, $\mathcal{A}_{12}$ defeats $\mathcal{A}_{5}$, and $\mathcal{A}_{12}$ is thus warranted in $\epistcompabs_\run$ and $\neg better(box_2,box_1)$ is justified in $\epistcompabs_\run$.
	
	Finally, note that the active arguments $\mathcal{A}_{1},$ $\mathcal{A}_{2},$ $\mathcal{A}_{3},$ $\mathcal{A}_{4},$ $\mathcal{A}_{7},$ $\mathcal{A}_{8},$ $\mathcal{A}_{9},$ $\mathcal{A}_{10},$  $\mathcal{A}_{13},$ $\mathcal{A}_{14},$ $\mathcal{A}_{15},$ $\mathcal{A}_{16}$ are not in conflict with any other active argument, and therefore, they are all warranted.
	Observe that one conclusion can be supported by more than one argument.
	For instance, $\mathcal{A}_{1},$ $\mathcal{A}_{7}$ and  $\mathcal{A}_{13}$ are three warranted arguments that support the same conclusion: $better(box_1,box_3)$.
	Then, from the warranted arguments mentioned above, the
	set of justified conclusions in $\epistcompabs_\run$
	is \{
	$better(box_1,box_3)$, 
	$better(box_2,box_3)$, 
	$better(box_1,box_2)$, 
	$\neg better(box_3,box_1)$, 
	$\neg better(box_3,box_2)$, 
	$\neg better(box_2,box_1)$
	\}.
\end{example}

We have shown in Example~\ref{exampleEpistemicComponent} that given a working set of arguments $W_\run$, the available evidence $E_\run$ will determine a set  $ \mathit{Active}_\run \subseteq W_\run$ with the current set of active arguments.
Then, the conflict relation $ \bowtie_\run$ and the preference function
$\mathfrak{pref}_\run$ are used for determining
the set  $\mathit{Warranted}_\run \subseteq \mathit{Active}_\run$  of the
current warranted arguments.
We will introduce next the decision component of our framework; this component will use the  justified  conclusions supported by arguments in $\mathit{Warranted}_\run$ to determine which alternatives will be selected.

\section{Accepting alternatives}
\label{sec:accepting}

We will formalize now the decision component \Dcomp\ of our proposed abstract decision framework \ADeF$_\lang$, and we will propose an algorithm for computing the selecting alternatives.
As we have mentioned, the decision component is a set of decision rules that will effectively implement the agent's decision making policy.
Decision rules~\cite{JETAI2014}, that will be used to decide among alternatives from a choice experiment $B \subseteq X$, are formalized in following definition.

\begin{definition}[Decision rule]\label{def:dec-rule}
	Let $X$ be a set of alternatives and  $B \subseteq X$ be a choice experiment $(B\not = \emptyset)$.
	A decision rule is denoted $(\desirule{D}{B}{P}{T})$, where
	$D \subseteq B$ represents the set of alternatives that this rule will select, $P \subseteq$ \lang\ represents preconditions for using this rule, and  $T \subseteq$ \lang\ represents the constraints that the rule has for its usage.
\end{definition}

A decision rule ``\desirule{D}{B}{P}{T}'' can be read as ``if all the preconditions included in $P$ hold and no constraint of the set $T$ holds then $D$ is the subset of alternatives from $B$ to be selected''.
Hence, $D$ will represent those alternatives that this rule decides to adopt from the choice experiment $B$  posed to the decision maker.
The following definition introduces our proposed decision component.
\begin{definition}[Decision component]
	\label{def:decision:component}
	Given a set of alternatives  $B \subseteq X$, and the schematic variables $W$, $Y$, and $Z$.
	The \textit{decision component} is the set  $\Gamma = \DRa \cup \DRb$,  where:
	\begin{itemize}
		\item[$\centerdot$] \DRa\ is a set of decision rules  obtained instantiating  the variables $W,Y,$ and $Z$  with constants from $B$ in the decision rule
		\[
		\desirule{\{W\}}{B}{\{better(W,Y)\}}{\{better(Z,W)\}}
		\]
		\item[$\centerdot$]  \DRb\ is a set of decision rules  obtained instantiating  the variables $W,Y,$ and $Z$  with constants from $B$ in the decision rule
		\[
		\desirule{\{W,Y\}}{B}{\{\neg better(W,Y),\neg better(Y,W)\}}{\{better(Z,W), better(Z,Y)\}}
		\]
	\end{itemize}
\end{definition}

On the one hand, a decision rule in the set $\DRa$ states that an alternative $W \in$~\ourB\ will be chosen, if $W$ is better than another alternative $Y$ and there is no better alternative $Z$ than $W$.
On the other hand, a decision rule in the set $\DRb$ states that two alternatives $W$ and $Y$ will be chosen when $W$ and $Y$ have the same attributes, and there is not other alternative $Z$ better than them.

The following definition states when a decision rule will be applicable with respect to a particular scenario which will be represented by the epistemic component \epistcomp.

\begin{definition}[Applicable decision rule]\label{def:rule-app}
	\noindent
	Let $B \subseteq X$  be a choice experiment and \epistcomp\ be an epistemic component. A decision rule (\desirule{D}{B}{P}{T})  is applicable with respect to \epistcomp,
	if every precondition in $P$ is justified in \epistcomp\ and every constraint in $T$ fails to be justified in \epistcomp.
\end{definition}

Then, to determine which alternatives will be selected from a choice experiment $B$ it is necessary to consider all the  decision rules from $\Gamma$ that are applicable with respect to the justified conclusions that can be inferred from the active arguments of \epistcomp.
The set of acceptable alternatives can be defined as follows.

\begin{definition}[Set of acceptable alternatives] \label{def:set-accept}
	Let $B \subseteq X$ be a set of alternatives posed to the agent and \ADeF\ be the agent's decision framework.
	Then, the set of acceptable alternatives $\alternatives{B}$ of the agent will be defined as follows:
	\center $D \subseteq \alternatives{B}$ iff
	$(\desirule{D}{B}{P}{T}) \in \Gamma$ is applicable with respect to \epistcomp.
\end{definition}

Clearly, if $B=\emptyset$ then no alternative is eligible, and
if $B$ is a singleton then there is no choice and the unique element of $B$ should be selected.
In Figure~\ref{fig.algorithm} we propose an algorithm for selecting acceptable alternatives from $B$ given a decision framework \ADeF.

\begin{figure}[ht]
	\noindent\makebox[\linewidth]{\rule{\textwidth}{0.8pt}}
	\begin{algorithm}[Compute acceptable alternatives]\ \\
		\label{algo:accept}
		\begin{small}
			
			\textsf{Input:}  a decision framework \ADeF\ and a set $B\subseteq X$.\\
			\textsf{Output:} a set $S \subseteq B$ \\
			
			$S=\emptyset$ \\
			\textsf{if} $B$ is a singleton \textsf{then} $S=B$ \\
			\textsf{else} \\
			\hspace*{5pt}\textsf{begin} \\
			\hspace*{10pt}\textsf{for each} $e \in B$ \textsf{do} \\
			\hspace*{20pt}\textsf{if} $(\desirule{\{e\}}{B}{\{better(e,Y)\}}{\{better(Z,e)\}})$
			is aplicable with respect to \epistcomp \\
			\hspace*{20pt}\textsf{then} $S=S\cup \{e\}$ \\
			\hspace*{10pt}\textsf{for each} $e_1, e_2 \in B$ \textsf{do}\\
			\hspace*{20pt}\textsf{if} $(\desirule{\{e_1, e_2\}}{B}{\{\neg better(e_1,e_2), \neg better(e_2,e_1)\}}{\{better(Z,e_1),better(Z,e_2)\}})$ \\
			\hspace*{20pt} is aplicable with respect to \epistcomp \\
			\hspace*{20pt}\textsf{then} $S=S\cup \{e_1,e_2\}$ \\
			\hspace*{5pt}\textsf{end}
		\end{small}
	\end{algorithm}
	\caption{Algorithm for computing acceptable alternatives}
	\noindent\makebox[\linewidth]{\rule{\textwidth}{0.8pt}}
	\label{fig.algorithm}
\end{figure}

\begin{remark}\label{remark:algo}
	Algorithm $\ref{algo:accept}$ implements a choice rule.
\end{remark}

Note that this algorithm implements a \textit{choice rule} as introduced in Section~\ref{sec:theory}.
That is because the algorithm assigns to each set of alternatives \ourB\ $\in \mathcal{B}$ a non-empty set that represents the alternatives that the decision maker \emph{might} choose when presented with the alternatives in \ourB.

\begin{example}
	Given the set of justified conclusions from
	\mbox{$\epistcompabs_\run=\langle E_\run, W_\run, \bowtie_\run, \mathfrak{pref}_\run \rangle$}
	of Example~\ref{exampleEpistemicComponent}:
	\{$better(box_1,box_3)$, 
	$better(box_2,box_3)$, 
	$better(box_1,box_2)$, \\
	$\neg better(box_3,box_2)$, 
	$\neg better(box_2,box_1)$\}.\\
	Consider now that  Algorithm~$\ref{algo:accept}$ is applied with $B =\{box_3, box_2, box_1\}$ and $\epistcompabs_\run$.
	Observe that if $e = box_3$ then the rule $(\desirule{\{e\}}{B}{\{better(e,Y)\}}{\{better(Z,e)\}})$
	with $\{Z,Y\} \subseteq B$  is not applicable with respect to $\epistcompabs_\run$ and therefore $box_3$ will not be selected.
	This is so because there is no justification indicating that $box_3$ is better than another box.
	Consider now $e = box_2$, then $box_2$ is not selected because
	there is a justification for $better(box_1,box_2)$.
	Nevertheless, when the algorithm considers
	$e = box_1$ then $box_1$ is selected $(S = \{box_1\})$
	because the rule
	$(\desirule{\{e\}}{B}{\{better(e,Y)\}}{\{better(Z,e)\}})$
	is applicable with respect to $\epistcompabs_\run$
	with $Y = box_2$ and $Z\in B$.
	Finally note that no decision rule in $\DRb$ of $\Gamma$
	is applicable with respect to $\epistcompabs_\run$.
	Therefore, the output of Algorithm~\ref{algo:accept} is $S = \{box_1\}$.
\end{example}

In Example~\ref{exampleEpistemicComponent}, the justified conclusions
of the epistemic component $\epistcompabs_\run$ were obtained considering a particular order among distinguished literals that was introduced in Example~\ref{ex.dist.lit.and.order}:\\[3pt]
$>_{\mathcal{C}_{\run}}=  \{(nearer\_robot,nearer\_store),$ $(nearer\_robot,smaller),$ \\
\hspace*{32pt}$(nearer\_store,smaller)\}$.\\[3pt]
We will show next that if a different order is considered then the justified conclusions can change.

\begin{example} 
	Consider again our running example (Figure~\ref{fig.outline}),  where $B =\{box_3, box_2, box_1\}$, now  with a new order of distinguished literals:\\[3pt]
	$>'_{\mathcal{C}_{\run}}=\{(nearer\_store,nearer\_robot),$ $(nearer\_robot,smaller),$\\ \hspace*{32pt}$(nearer\_store,smaller)\}$.\\[3pt]
	That is, in contrast to $>_{\mathcal{C}_{\run}}$,
	with the new order $>'_{\mathcal{C}_{\run}}$ the agent will prefer first boxes that are near to the store, and then boxes near to itself and then the smaller ones.
	With this new order $\mathfrak{pref}_\run(\mathcal{A}_{11}, \mathcal{A}_{6}) = \mathcal{A}_{6}$, and
	$\mathfrak{pref}_\run(\mathcal{A}_{5}, \mathcal{A}_{12}) = \mathcal{A}_{5}$.
	Then, the set of justified conclusions in  $\epistcompabs_\run$ is:\\
	\{$better(box_1,box_3)$, 
	$better(box_2,box_3)$, 
	$better(box_2,box_1)$, 
	\\
	$\neg better(box_3,box_1)$, 
	$\neg better(box_3,box_2)$, 
	$\neg better(box_1,box_2)$ \}.
	
	\noindent And then, with this set of justified conclusions the output of Algorithm~$\ref{algo:accept}$
	is $S = \{box_2\}$.
\end{example}
\begin{figure}[h]
	\centering
	\includegraphics[width=12cm]{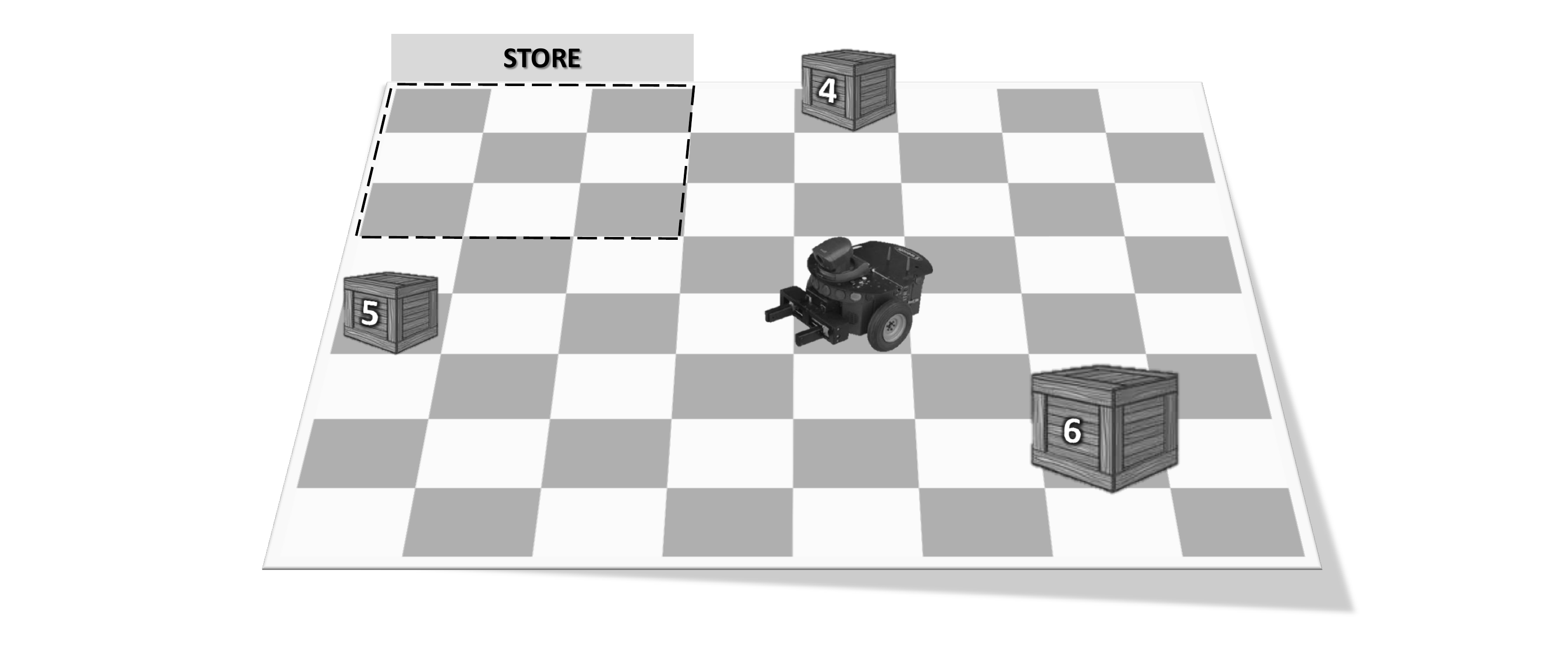}
	\caption{Two boxes with the same attribute values (Examples~\ref{example_sameprop} and~\ref{ex.other.order}).}
	\label{fig.sameprop}
\end{figure}

\newcommand{\two}{2}
\begin{example}\label{example_sameprop} 
	Given the new scenario depicted in  Figure~\ref{fig.sameprop}  where $\SAP=\{box_4,box_5,$ $box_6\}$.
	We will use the following set of distinguished literals
	$\mathcal{C}_{2}$ $=$ $\{nearer\_robot,$ $nearer\_store,$ $smaller\}$ and the order
	$>_{\mathcal{C}_{2}}$ = $\{(nearer\_store,$ $nearer\_robot),$ $(nearer\_robot,$ $smaller),$ $(nearer\_store,$ $smaller)\}$.
	Note that $>_{\mathcal{C}_{2}}$ prefers boxes near the store, then boxes close to the robot, and finally small boxes.
	The epistemic component $\epistcompabs_\two=\langle
	E_\two, W_\two, \bowtie_\two, \mathfrak{pref}_\two \rangle$
	for the scenario of Figure~\ref{fig.sameprop} will have the following set of evidence:
	\[ E_\two=\left\{
	\begin{tabular}{l l}
	$smaller(box_4,box_6),$&$nearer\_robot(box_6,box_4),$\\
	$smaller(box_5,box_6),$&$nearer\_robot(box_6,box_5),$\\
	$nearer\_store(box_4,box_6),$&$same\_att(box_4,box_5)$ \\
	$nearer\_store(box_5,box_6),$
	\end{tabular}
	\right\}
	\]
	\begin{figure}[]
		\begin{center}
			\includegraphics[width=114mm]{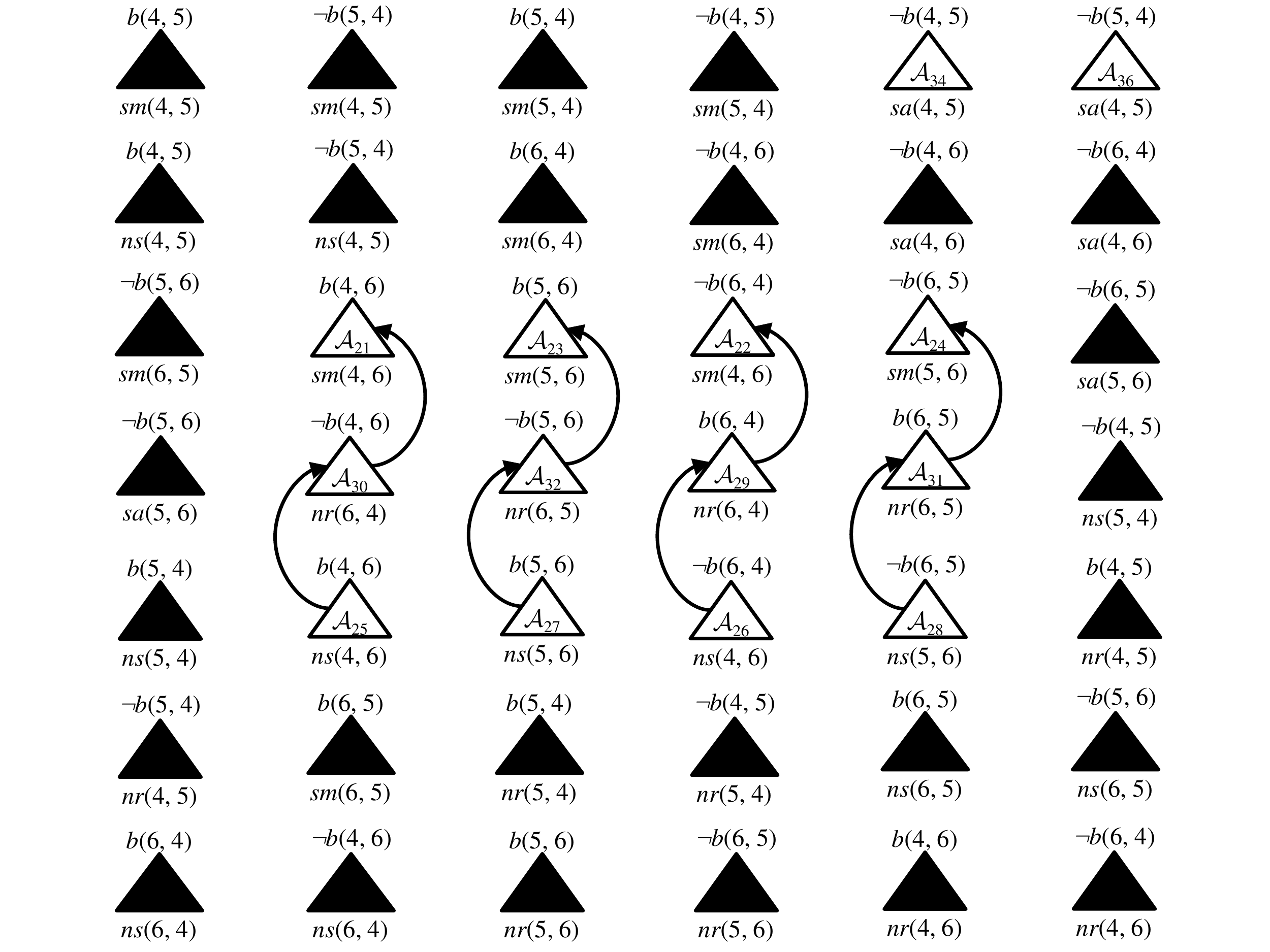}
			\caption{Active and inactive arguments from the working set
				$W_2$ of Example~\ref{example_sameprop}.} \label{fig:W2}
		\end{center}
	\end{figure}
	The working set $W_\two$ of $\epistcompabs_\two$ is depicted in Figure~\ref{fig:W2}.
	There are fourteen active arguments (white triangles) with respect to $E_\two$; as introduced above, arrows depict the defeat relation.
	The details of active arguments are showed below.
	\begin{center}
		\begin{tabular}{l}
			$\mathcal{A}_{21} =\langle \{smaller(box_4,box_6)\}, better(box_4,box_6)\rangle$,\\
			$\mathcal{A}_{23}=\langle \{smaller(box_5,box_6)\}, better(box_5,box_6)\rangle$,\\
			$\mathcal{A}_{25}=\langle \{nearer\_store(box_4,box_6)\}, better(box_4,box_6)\rangle$,\\
			$\mathcal{A}_{27}=\langle \{nearer\_store(box_5,box_6)\}, better(box_5,box_6)\rangle$,\\
			$\mathcal{A}_{29}=\langle \{nearer\_robot(box_6,box_4)\}, better(box_6,box_4)\rangle$,\\
			$\mathcal{A}_{31}=\langle \{nearer\_robot(box_6,box_5)\}, better(box_6,box_5)\rangle$,\\
			$\mathcal{A}_{22}=\langle \{smaller(box_4,box_6)\}, \neg better(box_6,box_4)\rangle$,\\
			$\mathcal{A}_{24}=\langle \{smaller(box_5,box_6)\}, \neg better(box_6,box_5)\rangle$,\\
			$\mathcal{A}_{26}=\langle \{nearer\_store(box_4,box_6)\}, \neg better(box_6,box_4)\rangle$,\\
			$\mathcal{A}_{28}=\langle \{nearer\_store(box_5,box_6)\}, \neg better(box_6,box_5)\rangle$,\\
			$\mathcal{A}_{30}=\langle \{nearer\_robot(box_6,box_4)\}, \neg better(box_4,box_6)\rangle$,\\
			$\mathcal{A}_{32}=\langle \{nearer\_robot(box_6,box_5)\}, \neg better(box_5,box_6)\rangle$,\\
			$\mathcal{A}_{34}=\langle \{same\_att(box_4,box_5)\}, \neg better(box_4,box_5)\rangle$,\\
			$\mathcal{A}_{36}=\langle \{same\_att(box_4,box_5)\}, \neg better(box_5,box_4)\rangle$
		\end{tabular}
	\end{center}
	Note that, $ \bowtie_{\two} = \{$
	$(\mathcal{A}_{21},\mathcal{A}_{30})$,
	$(\mathcal{A}_{25},\mathcal{A}_{30})$,
	$(\mathcal{A}_{23},\mathcal{A}_{32})$,
	$(\mathcal{A}_{27},\mathcal{A}_{32})$,
	$(\mathcal{A}_{29},\mathcal{A}_{22})$,
	$(\mathcal{A}_{29},\mathcal{A}_{26})$,
	$(\mathcal{A}_{31},\mathcal{A}_{24})$,
	$(\mathcal{A}_{31},\mathcal{A}_{28})$
	$\}$
	are the conflicts between active arguments.
	Note that active arguments $\mathcal{A}_{34}$ and $\mathcal{A}_{36}$ are not in conflict with any other argument.
	Considering the order $>_{\mathcal{C}_{\run}}$ then,
	\begin{tabbing}
		\hspace*{60pt}\=$\mathfrak{pref}_\run(\mathcal{A}_{21}, \mathcal{A}_{30}) = \mathcal{A}_{30}$, $\mathfrak{pref}_\run(\mathcal{A}_{25}, \mathcal{A}_{30}) = \mathcal{A}_{25}$,\\
		\>$\mathfrak{pref}_\run(\mathcal{A}_{23}, \mathcal{A}_{32}) = \mathcal{A}_{32}$, $\mathfrak{pref}_\run(\mathcal{A}_{27}, \mathcal{A}_{32}) = \mathcal{A}_{27}$,\\
		\>$\mathfrak{pref}_\run(\mathcal{A}_{29}, \mathcal{A}_{22}) = \mathcal{A}_{29}$, $\mathfrak{pref}_\run(\mathcal{A}_{29}, \mathcal{A}_{26}) = \mathcal{A}_{26}$,\\
		\>$\mathfrak{pref}_\run(\mathcal{A}_{31}, \mathcal{A}_{24}) = \mathcal{A}_{31}$, $\mathfrak{pref}_\run(\mathcal{A}_{31}, \mathcal{A}_{28}) = \mathcal{A}_{28}$.
	\end{tabbing}
	Since $\mathfrak{pref}_\run(\mathcal{A}_{25}, \mathcal{A}_{30}) = \mathcal{A}_{25}$,
	then   $\mathcal{A}_{25}$ is not defeated and hence,
	$\mathcal{A}_{25}$ is a warranted argument in  $\epistcompabs_\two$.
	Note that $\mathcal{A}_{30}$ is defeated and therefore is not a warranted argument.
	Finally, note that $\mathcal{A}_{21}$ is  defeated by $\mathcal{A}_{30}$,
	and  $\mathcal{A}_{30}$ is in turn defeated by $\mathcal{A}_{25}$, then $\mathcal{A}_{21}$ is a warranted argument (for more details see Definition~\ref{def:warrant:nico} at the Appendix).
	Thus, the set of warranted arguments in  $\epistcompabs_\two$ is
	\{$\mathcal{A}_{25}$, $\mathcal{A}_{21}$,
	$\mathcal{A}_{22}$, $\mathcal{A}_{27}$, $\mathcal{A}_{34}$, $\mathcal{A}_{36}$, $\mathcal{A}_{23}$, $\mathcal{A}_{26}$, $\mathcal{A}_{24}$, $\mathcal{A}_{28}$\}.
	Hence, the set of justified conclusions in  $\epistcompabs_\two$ is\\
	\{$better(box_4,box_6)$, 
	$better(box_5,box_6)$, 
	$\neg better(box_4,box_5)$, 
	\\$\neg better(box_5,box_4)$,  
	$\neg better(box_6,box_4)$, 
	$\neg better(box_6,box_5)$\}. 
	
	And then, with this set of justified conclusions the output of Algorithm~$\ref{algo:accept}$  is $S = \{box_4, box_5\}$.
\end{example}

In Example~\ref{example_sameprop}, the justified conclusions of the epistemic component $\epistcompabs_\two$ were obtained considering a particular order among distinguished literals.
When considering different orders imposed on the distinguished literals it is possible that the set of justified conclusions could change as we will show next.

\begin{figure}[h]
	\begin{center}
		\includegraphics[width=75mm]{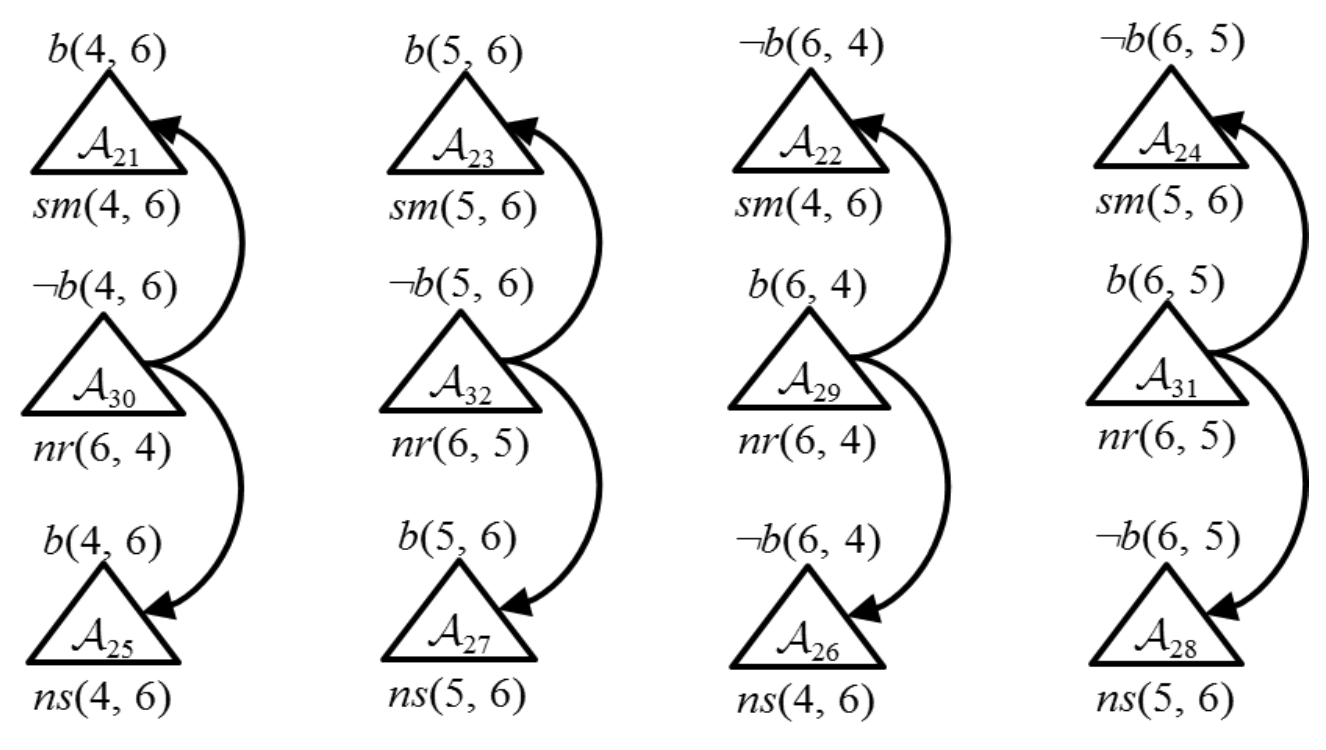}
		\caption{Defeat between conflictive arguments of Example~\ref{ex.other.order}.} \label{fig:defeat}
	\end{center}
\end{figure}

\begin{example}\label{ex.other.order} 
	Consider again the scenario depicted in Figure~\ref{fig.sameprop},
	with the set of alternatives $B =\{box_4, box_5, box_6\}$ but the distinguished literals with a different order:\\[3pt]
	\hspace*{24pt}$>'_{\mathcal{C}_{2}}=\{(nearer\_robot,nearer\_store),$ $(nearer\_robot,smaller),$\\ \hspace*{56pt}$(nearer\_store,smaller)\}$.\\[3pt]
	That is, in contrast to $>_{\mathcal{C}_{2}}$, with the order
	$>'_{\mathcal{C}_{2}}$ the agent will prefer first boxes that are near to itself,  then boxes close to the store and finally smaller boxes.
	Note that the active arguments are the same of Figure~\ref{fig:W2} but
	the preference between conflictive arguments change (see Figure~\ref{fig:defeat}).
	Then, with $>'_{\mathcal{C}_{2}}$ we have:
	\begin{tabbing}
		\hspace*{60pt}\=$\mathfrak{pref}_\run(\mathcal{A}_{21}, \mathcal{A}_{30}) = \mathcal{A}_{30}$, $\mathfrak{pref}_\run(\mathcal{A}_{25}, \mathcal{A}_{30}) = \mathcal{A}_{30}$,\\
		\>$\mathfrak{pref}_\run(\mathcal{A}_{23}, \mathcal{A}_{32}) = \mathcal{A}_{32}$, $\mathfrak{pref}_\run(\mathcal{A}_{27}, \mathcal{A}_{32}) = \mathcal{A}_{32}$,\\
		\>$\mathfrak{pref}_\run(\mathcal{A}_{29}, \mathcal{A}_{22}) = \mathcal{A}_{29}$, $\mathfrak{pref}_\run(\mathcal{A}_{29}, \mathcal{A}_{26}) = \mathcal{A}_{29}$,\\
		\>$\mathfrak{pref}_\run(\mathcal{A}_{31}, \mathcal{A}_{24}) = \mathcal{A}_{31}$, $\mathfrak{pref}_\run(\mathcal{A}_{31}, \mathcal{A}_{28}) = \mathcal{A}_{31}$.
	\end{tabbing}
	Hence, the set of justified conclusions is\\
	\{$better(box_6,box_4)$, 
	$better(box_6,box_5)$, 
	$\neg better(box_4,box_5)$, 
	\\$\neg better(box_5,box_4)$, 
	$\neg better(box_4,box_6)$ 
	$\neg better(box_5,box_6)$\}. 
	\\And with this set of justified conclusions, since $box_6$ is nearer to the robot, the output of Algorithm~\ref{algo:accept}  is $S = \{box_6\}$.
\end{example}

The precedent examples were designed to show the behavior of our proposal in some selected special cases.
Nevertheless, our framework is conceived to work in a dynamic environment where the attribute values of the alternatives can change.
Every time the decision maker perceives a change in the environment this event will
change the available evidence.
In a DAF, when the set of evidence $E$ changes, the set of active arguments of $W$ can change, and hence, warranted arguments and justified conclusions can change.
Therefore, a change in the set $E$ can affect the set of acceptable alternatives of the agent introducing changes in it;
thus, one feature of our formalization is that there is no need to have an explicit formalization of time.
For instance, if in the scenario depicted in Figure~\ref{fig.sameprop}, $box_5$ is moved next to the robot, this change in the evidence set will change the set of acceptable alternatives, and the algorithm will choose $box_5$.
As a further example, if in the scenario depicted in Figure~\ref{fig.sameprop}  $box_6$ is moved down to the leftmost corner, the algorithm will choose \{$box_4$,$box_5$\}.

Next, in Section \ref{sec:formal}, we will present results that show what happens when two alternatives have different attribute values and what happens when they have the same attribute values.
These propositions will be used to relate the choice behavior exhibited by the decision framework with the choice behavior from the approaches based on preferences and choice rules of classical decision theory.

\section{Formal comparison with Classical Decision Theory}
\label{sec:formal}

Hereinafter, we will formally relate our proposal to Classical Decision Theory.
The first two propositions show what happens when two alternatives have different properties and what occurs when they have the same attributes (\ie~they have the same attribute values for each preference criterion);
these propositions will be used to formalize the choice behavior of the \emph{abstract decision framework} introduced in Definition~\ref{def:abst:df}.
The formalization comprises Lemma~\ref{lem:sdast:rat:abs}, and Theorems~\ref{teo:pref-rel:abs} and~\ref{teo:rel:pr:cs:fr:abs} which relate the choice behavior exhibited by the abstract decision framework with the choice behavior from the approaches of classical decision theory which are based on preferences and choice rules.
For some of the proofs in this section we will use some DAF's concepts and notation~\cite{RotsteinMGS10};
to facilitate reading, we have included part of this formalism in the Appendix, in particular, the corresponding definitions for $args(\cdot)$, \emph{cl}$(\cdot)$, $\inti{\cdot}$, \emph{pr}$(\cdot)$, \emph{dialectical tree} $\mathcal{T}_{\mathbf{F}}(\cdot)$, \emph{argumentation line} $\lambda$, and skeptical marking function $\mathfrak{m}_e(\cdot)$, which are used below, can be found there.

\begin{proposition}
	\label{prop:warrant:defea:abs} Let $\epistcompabs = \langle E, W,\bowtie, \mathfrak{pref} \rangle$ be an abstract epistemic component, \SAP\ be the set with all the possible alternatives, and let $x,y \in X$ be two alternatives with different properties.
	Then, either there exist an argumental structure $\Sigma \in \domEA$ that is warranted from \epistcompabs\ and $cl(\Sigma)= $ \mbox{$better(x,y)$} or there exist an argumental structure $\Sigma^\prime \in \domEA$ that is warranted in \epistcompabs\ and $cl(\Sigma^\prime)= $\mbox{$\,\neg better(x,y)$}.
	\begin{proof}
		By hypothesis, alternatives $x$ and $y$ have different properties.
		This means that their attribute values differ in at least one preference criterion.
		
		If $p_j$ is just the one criterion where $x$ and $y$ differ, either $c_j(x,y)$ or $c_j(y,x)$ will be an element of evidence set $E$ (Definition~\ref{def:epist:comp:abs}).
		If \mbox{$c_j(x,y) \in E$}, then (by Definition~\ref{def:epist:comp:abs}) there exists a primitive argumental structure \EA{j} such that
		$\args{\EA{j}}=\{\Argu_{j}\}$ and
		\inti{\Argu_{j}}~$=$~\intd{\facto{c_j(x,y)}}{\facto{better(x,y)}},
		which is active with respect to $E$.
		Besides, an active dialectical tree \ArbD{\EA{j}} will be built for \EA{j} that will consist of only one
		argumentation line $\lambda = [$\EA{j}$]$, such that it trivially holds that
		$\mathfrak{m}_e(\Sigma_j,\lambda,\mathbb{T}_{{\scriptscriptstyle\mathcal{K}_{\mathcal{A}}}}(\Sigma_j))=\mathbf{U}$.
		
		Conversely, if $c_j(y,x) \in E$, then a primitive argumental structure $\Sigma^\prime_j$ will exist such that
		$\args{\Sigma^\prime_j}=\{\Argu^\prime_{j}\}$ and
		\inti{\Argu^\prime_{j}}~$=$~\intd{\facto{c_j(y,x)}}{\facto{\neg better(x,y)}}, which is active with respect to $E$.
		Moreover, an active dialectical tree
		\ArbD{$\Sigma^\prime_j$} will be built for $\Sigma^\prime_j$ that will consist of only one
		argumentation line
		$\lambda^\prime = [\Sigma^\prime_j]$, such that it also holds that the marking function
		$\mathfrak{m}_e(\Sigma^\prime_j,\lambda^\prime,\mathbb{T}_{{\scriptscriptstyle
				\mathcal{K}_{\mathcal{A}}}}(\Sigma^\prime_j))=\mathbf{U}$.
		
		Alternatively, if more than one preference criteria exist where $x$ and $y$ differ, two cases may hold:
		\begin{itemize}[topsep=2pt]\itemsep 2pt\parskip 0pt
			\item[(\emph{i})] that one alternative ($x$ or $y$) is better than the other one with respect to all the preference criteria where they differ, or
			\item[(\emph{ii})] there are at least two criteria according to which one alternative is considered better than the other one and vice versa.
		\end{itemize}

		Case (\emph{i}) is a generalization of the case analyzed above, where $x$ and $y$ only differ in exactly one preference criterion; and the result of the dialectical analysis is the same. Only one primitive argumental structure will be warranted from \epistcompabs\ supporting either $better(x,y)$ or $\neg
		better(x,y)$, depending on which alternative is deemed better wrt. all preference criteria.
		
		In case (\emph{ii}), two types of primitive argumental structures are distinguished; namely those supporting $better(x,y)$ and the ones supporting $\neg better(x,y)$.
		These argumental structures will be organized in an acceptable argumentation line $\lambda$ that can be decomposed in a set of argumental structures \emph{pro}
		$\lambda^+=[\Sigma_1,\ldots]$ (those supporting conclusion
		$better(x,y)$) and another set \emph{cons} $\lambda^-=[\Sigma_2,\ldots]$ (those
		supporting conclusion $\neg better(x,y)$). If $>_{\mathcal{C}}$
		is a strict total order, given two argumental structures
		\EA{i}~$\in \lambda^+$ and \EA{j}~$\in \lambda^-$,
		$\mathfrak{pref}(\Sigma_i,\Sigma_j)=\Sigma_i$ or
		$\mathfrak{pref}(\Sigma_i,\Sigma_j)=\Sigma_j$.
		This is due to the fact that in this case $\mathfrak{pref}(\Sigma_i,\Sigma_j)$ will never be $\epsilon$, since alternatives $x$ and $y$ have different properties.

		In this way, for $\Sigma_1$ there will be just one active dialectical tree \ArbD{$\Sigma_1$} containing only one argumentation line
		$\lambda$, such that
		$\mathfrak{m}_e(\Sigma_1,\lambda,\mathbb{T}_{{\scriptscriptstyle
				\mathcal{K}_{\mathcal{A}}}}(\Sigma_1))=\mathbf{U}$ if $|\lambda^+| >
		|\lambda^-|$, and
		$\mathfrak{m}_e(\Sigma_1,\lambda,\mathbb{T}_{{\scriptscriptstyle
				\mathcal{K}_{\mathcal{A}}}}(\Sigma_1))=\mathbf{D}$ otherwise.
	\end{proof}
\end{proposition}

\begin{corollary}
	\label{coro:warrant:defea:abs} Let $\epistcompabs = \langle E, W,
	\bowtie, \mathfrak{pref} \rangle$ be an abstract epistemic component, \SAP\ be the set with all the possible alternatives, and let $x,y \in X$ be two alternatives with different properties. If
	$>_{\mathcal{C}}$ is a strict total order and either conclusion $better(x,y)$ or $\neg better(x,y)$ is warranted in
	\epistcompabs, then either $\neg better(y,x)$ or $better(y,x)$
	is also warranted in \epistcompabs, respectively.
\end{corollary}

Proposition~\ref{prop:warrant:defea:abs} states that either $better(x,y)$ or $\neg better(x,y)$ will be warranted by the epistemic component when two alternatives $x,y \in X$ with different properties are compared;
this means that it is always possible to decide which alternative is preferred.
In Definition~\ref{def:epist:comp:abs}, the features of arguments belonging to the working set are introduced, and hence how a piece of evidence $c(x,y)$ activates arguments supporting $better(x,y)$ or $\neg better(y,x)$. Corollary~\ref{coro:warrant:defea:abs} states that when $better(x,y)$ or $\neg better(x,y)$ is warranted, it will also be the case that $\neg better(y,x)$ or $better(y,x)$ is also warranted; this corollary, complements the above-mentioned proposition in the symmetry of comparing alternative $x$ versus $y$, or the opposite, depending on how they are presented in the facts available in the evidence set.

In a similar manner, Proposition~\ref{prop:warrant:strict:abs} below states that when two alternatives $x,y \in X$ have the same attributes, comparing them in the epistemic component will result in warranting conclusions $\neg better(x,y)$ and $\neg better(y,x)$, stating the indifferent preference among them.

\begin{proposition}
	\label{prop:warrant:strict:abs} Let $\epistcompabs=\langle E, W,\bowtie, \mathfrak{pref} \rangle$ be an abstract epistemic component, \SAP\ be the set with all the possible alternatives, and let $x,y \in X$ be two alternatives with the same attributes.
	If $>_{\mathcal{C}}$ is a total strict order, then there exist two argumental structures
	$\Sigma, \Sigma^\prime \in \mathfrak{str}_{{\scriptscriptstyle \mathcal{K}_{\mathcal{A}}}}$
	warranted in \epistcompabs\ such that $cl(\Sigma)=\neg better(x,y)$ and $cl(\Sigma^\prime)=\neg better(y,x)$.
	\begin{proof}
		By hypothesis, alternatives $x$ and $y$ have the same attributes, therefore a fact $same\_att(x,y)$ will belong to
		evidence set $E$ (Definition~\ref{def:epist:comp:abs}).
		Thus, by the definition of working set $W$ (Definition~\ref{def:epist:comp:abs}) two primitive structures \EA{} and $\Sigma^\prime$ exist such that set of arguments in \EA{} and $\Sigma^\prime$ are respectively
		\args{\EA{}}~$=\{$\Argu$\}$,
		\args{\Sigma^\prime}~$=\{$\Argu$^\prime\}$,\footnote{See Appendix for notation.} with the corresponding interfaces
		$$\inti{\Argu}~=~\intd{\facto{same\_att(x,y)}}{\facto{\neg better(x,y)}}$$
		\hspace*{16pt} and
		$$\inti{\Argu^\prime}~=~\intd{\facto{same\_att(x,y)}}{\facto{\neg better(y,x)}},$$
		and which are active with respect to $E$.
		Moreover, two active dialectical trees, \ArbD{\EA{}} and \ArbD{$\Sigma^\prime$}, will be built each one consisting of only one argumentation line $\lambda = [\EA{}]$ and $\lambda^\prime = [\Sigma^\prime]$, respectively, such that it trivially holds that
		$\mathfrak{m}_e(\Sigma,\lambda,\mathbb{T}_{{\scriptscriptstyle
				\mathcal{K}_{\mathcal{A}}}}(\Sigma))=\mathbf{U}$ and
		$\mathfrak{m}_e(\Sigma^\prime,\lambda^\prime,\mathbb{T}_{{\scriptscriptstyle
				\mathcal{K}_{\mathcal{A}}}}(\Sigma^\prime))=\mathbf{U}$.
	\end{proof}
\end{proposition}

The results stated in Propositions~\ref{prop:warrant:defea:abs}~and~\ref{prop:warrant:strict:abs} provide formal support for Lemma~\ref{lem:sdast:rat:abs}, which states that given two alternatives $x$ and $y$ it will always be possible to compare them in the epistemic component, whether to express indifference or a preference for one of the alternatives (completeness).
Besides, this lemma also states that the epistemic component satisfies the property of transitivity when all alternatives are pairwise compared.

\begin{lemma}
	\label{lem:sdast:rat:abs} Let $\epistcompabs=\langle E, W, \bowtie,
	\mathfrak{pref} \rangle$ be an abstract epistemic component. If
	$>_{\mathcal{C}}$ is a strict total order, then
	\epistcompabs\ implements\hspace{2pt}\footnote{This term means that given two alternatives $x$ and $y$ it will always be possible to compare them. In fact, there will always be an active argumental structure $\Sigma$ warranted in \epistcompabs\
		such that $cl(\Sigma)=Z$ with $Z \in \{better(x,y),better(y,x),\neg
		better(x,y),\neg better(y,x)\}$, depending on the properties of the alternatives.} a rational preference relation \ourRelPref.
	\begin{proof}
		\label{demost:lem:sdast:rat:abs}
		By Definition~\ref{def:rational}, given two alternatives $x,y \in X$, then $x \ourRelPref y \Leftrightarrow x
		\ourRelPrefEst y \vee x \ourRelPrefInd y$.
		From Definition~\ref{def:epist:comp:abs}, the evidence set $E$ will contain all the facts of the kind $c(x,y)$ relating alternatives $x$ and $y$ (that belong to the choice experiment posed to the decision maker) with respect to all the preference criteria referred by the distinguished literals  in $\mathcal{C}$.
		In this way, by Proposition~\ref{prop:warrant:defea:abs} and Corollary~\ref{coro:warrant:defea:abs} it will always be possible to activate a primitive argumental structure $\Sigma$, warranted in \epistcompabs, such that $cl(\Sigma)=Z$ with $Z \in \{better(x,y),better(y,x),\neg better(x,y),\neg better(y,x)\}$, thus stating a strong preference ($\succ$) in favor of one of the alternatives.
		Conversely, if alternatives $x$ and $y$ have the same attributes, by Proposition~\ref{prop:warrant:strict:abs}, conclusions $\neg better(x,y)$ and $\neg better(y,x)$ will be warranted by two active primitive argumental structures, which are warranted in \epistcompabs, thus stating the indifference ($\sim$) on the preference between the alternatives.
		Therefore, the completeness property is satisfied.
		
		In order to check transitivity property, we must prove that for all $x, y, z \in X$, if $x \ourRelPref y$ and $y \ourRelPref z$, then $x \ourRelPref z$.
		If $x \ourRelPref y$ is satisfied, this means that:
		\begin{itemize}[topsep=4pt]\itemsep 2pt\parskip 0pt
			\item[(\emph{i})] $x \succ y$: by Proposition~\ref{prop:warrant:defea:abs}, the conclusion $better(x,y)$ is warranted given that a primitive argumental structure $\Sigma_i$ exists (active wrt. $E$), which is warranted in \epistcompabs\ such that $cl(\Sigma_i)=better(x,y)$, or
			\item[(\emph{ii})] $x \sim y$: by Proposition~\ref{prop:warrant:strict:abs},  the conclusions $\neg better(x,y)$ and
			$\neg better(y,x)$ are warranted since two active primitive argumental structures $\Sigma_j$ and
			$\Sigma_{j^\prime}$ exist and they are warranted in \epistcompabs\ and
			$cl(\Sigma_j)=\neg better(x,y)$ and $cl(\Sigma_{j^\prime})=\neg
			better(y,x)$, respectively.
		\end{itemize}
		Similarly, if $y \ourRelPref z$ holds, this means that:
		\begin{itemize}[topsep=4pt]\itemsep 2pt\parskip 0pt
			\item[(\emph{iii})] $y \succ z$: by Proposition~\ref{prop:warrant:defea:abs}, the conclusion $better(y,z)$ is warranted given that a primitive argumental structure $\Sigma_k$ exists (active wrt. $E$), which is warranted in
			\epistcompabs\ such that $cl(\Sigma_k)=better(y,z)$, or
			\item[(\emph{iv})] $y \sim z$: by Proposition~\ref{prop:warrant:strict:abs}, the conclusions $\neg better(y,z)$ and
			$\neg better(z,y)$ are warranted since two active primitive argumental structures $\Sigma_l$ and
			$\Sigma_{l^\prime}$ exist and they are warranted in \epistcompabs\ and
			$cl(\Sigma_l)=\neg better(y,z)$ and $cl(\Sigma_{l^\prime})=\neg
			better(z,y)$, respectively.
		\end{itemize}
		If case (\emph{i}) holds, there exists an active dialectical tree
		$\mathbb{T}_{{\scriptscriptstyle
				\mathcal{K}_{\mathcal{A}}}}(\Sigma_i)$ such that
		$\mathfrak{m}_e(\Sigma_i,\lambda_i,\mathbb{T}_{{\scriptscriptstyle
				\mathcal{K}_{\mathcal{A}}}}(\Sigma_i))=\mathbf{U}$ where
		$\lambda_i=[\Sigma_i, \ldots, \Sigma_n]$.\footnote{It is worth noting that in the case of $n=i$, it results in the argumentation line $\lambda_i=[\Sigma_i]$.}
		Given that $\Sigma_n$ is the last argumental structure in the argumentation line, this implies that a distinguished literal $c_n$ exists such that $pr(args(\Sigma_n))=\{c_n(x,y)\}$ and no literal $c^\prime \in \mathcal{C}$ exists with higher priority than $c_n$ such that $c^\prime(y,x) \in E$.
		If $>_{\mathcal{C}}$ is a strict total order, from Definition~\ref{def:epist:comp:abs} it can be stated that $y$ cannot be better than $x$ with respect to any preference criterion with higher priority than $c_n$; if it could, then $\Sigma_n$ would not be the last argumental structure in the argumentation line since there will be another structure attacking it and that is based on this supposed criterion.
		
		Likewise, if case (\emph{iii}) holds, then an active dialectical tree  $\mathbb{T}_{{\scriptscriptstyle
				\mathcal{K}_{\mathcal{A}}}}(\Sigma_k)$ exists such that $\mathfrak{m}_e(\Sigma_k,\lambda_k,\mathbb{T}_{{\scriptscriptstyle\mathcal{K}_{\mathcal{A}}}}(\Sigma_k))=\mathbf{U}$ where $\lambda_k=[\Sigma_k, \ldots, \Sigma_m]$.\footnote{It is worth noting that in the case of $m=k$  it results in the argumentation line  $\lambda_k=[\Sigma_k]$.}
		Given that $\Sigma_m$ is the last argumental structure in the argumentation line, this implies that a distinguished literal $c_m$ exists such that $pr(args(\Sigma_m))=\{c_m(y,z)\}$  and no literal $c^\prime \in \mathcal{C}$ exists with higher priority than $c_m$ such that $c^\prime(z,y) \in E$.
		Again, if $>_{\mathcal{C}}$ is a strict total order, based on Definition~\ref{def:epist:comp:abs}, it can be stated that $z$ is not better than $y$  with respect to any preference criterion with higher priority than  $c_m$.
		As mentioned in the analysis of case (\emph{i}), this is due to the fact that if another criterion with higher priority would exist, for which $z$ is preferred to $y$, $\Sigma_m$ would not be the last structure in the argumentation line since there would be another one attacking it based on this supposed criterion.
		
		When considering cases (\emph{i}) and (\emph{iii}) together, it remains to compare $c_n$ with $c_m$ to determine whether $better(x,z)$ can be warranted; it can be the case that $c_n=c_m$, or $c_n > c_m$, or $c_m > c_n$.
		Independently of how these criteria are related each other, it will always be possible to guarantee that $z$ is not better than $x$ with respect to preference criteria of higher or equal priority than $c^{\prime \prime}=\max(c_n,c_m)$.
		
		In this way, based on Definition~\ref{def:epist:comp:abs}, we know that $c^{\prime
			\prime}(x,z) \in E$ when $x$ and $z$ belong to the same choice experiment, and consequently, a primitive argumental structure $\Sigma^{\prime \prime}$ will be activated such that
		$cl(\Sigma^{\prime \prime})=better(x,z)$, $pr(args(\Sigma^{\prime
			\prime}))=\{c^{\prime \prime}(x,z)\}$ and it will be warranted in
		\epistcompabs. This is due to the fact that an active dialectical tree
		$\mathbb{T}_{{\scriptscriptstyle
				\mathcal{K}_{\mathcal{A}}}}(\Sigma)$ will exist such that
		$\mathfrak{m}_e(\Sigma,\lambda,\mathbb{T}_{{\scriptscriptstyle
				\mathcal{K}_{\mathcal{A}}}}(\Sigma))=\mathbf{U}$ where
		$\lambda=[\Sigma, \ldots, \Sigma^{\prime \prime}]$.\footnote{Note that it could be the case of $\lambda=[\Sigma^{\prime \prime}]$.}
		
		If case (\emph{iv}) holds, this means that $y$ and $z$ have the same attributes. Hence, if we consider this case together with case (\emph{i}), it will be possible to build a dialectical tree analogous to the one built for case (\emph{i}), but warranting the conclusion $better(x,z)$.
		
		If case (\emph{ii}) holds, this means that $x$ and $y$ have the same attributes.
		In this way, if we consider this case together with case (\emph{iii}), it will be possible to build a dialectical tree analogous to the one built for case (\emph{iii}), but warranting the conclusion $better(x,z)$.
		
		Finally, if case (\emph{ii}) and (\emph{iv}) are considered together, by Proposition~\ref{prop:warrant:strict:abs} it can be stated that the conclusions $\neg better(x,z)$ and $\neg better(z,x)$ will be warranted since two active primitive argumental structures $\Sigma$ and $\Sigma^\prime$ exist which are warranted in
		\epistcompabs\ and $cl(\Sigma)=\neg better(x,z)$ and $cl(\Sigma^\prime)=\neg better(z,x)$, respectively.
		
		All in all, from the case analysis carried out above, it can be concluded that $x \ourRelPrefEst z
		\vee x \ourRelPrefInd z$, and hence $x \ourRelPref z$.
		Therefore, the completeness and transitivity properties are both satisfied.
	\end{proof}
\end{lemma}
The previous results (Lemma~\ref{lem:sdast:rat:abs}, Propositions~\ref{prop:warrant:defea:abs}~and~\ref{prop:warrant:strict:abs}) are aimed at formally characterizing properties of the epistemic component.
As described in Section~\ref{sec:accepting}, the decision component of the framework consists of a set of decision rules that will effectively implement the agent's decision making policy.
These rules use warranted information from the epistemic component to compare all the alternatives belonging to the choice experiment posed to the agent. The choice behavior of the decision framework is the result of this interaction between decision rules and the epistemic component; in particular, the following theorem establishes the coincidence of the choices made by our framework and the preference-based approach presented in Section~\ref{sec:theory}.

\begin{theorem}
	\label{teo:pref-rel:abs}
	Let \ADeF$_\lang$ be an abstract decision framework, where
	$\epistcompabs$ = $\langle E, W, \bowtie, \mathfrak{pref}
	\rangle$. Given a choice experiment \ourB~$\subseteq$~\SAP\
	posed to the agent, if $>_{\mathcal{C}}$ is a strict total order among elements of $\mathcal{C}$, the choice behavior of \ADeF$_\lang$ coincides
	with the optimum one of a rational preference relation.
	
	\begin{proof}
		As stated in Section~\ref{sec:theory}, an individual having a rational preference relation \ourRelPref\ on \SAP, will choose any element of the set \MPAB\ when facing a choice experiment \ourB\ $\subseteq$ \SAP; this is due to the fact of her preference-maximizing behavior.
		Besides, by Definition~\ref{def:set-accept}, $\alternatives{B}=\bigcup_{i=1}^{n}D_i$, where each set $D_i$
		contains the alternatives chosen by the applicable decision rule $i$. By Definition~\ref{def:rule-app}, a
		decision rule can be applied if its preconditions are warranted in \epistcompabs, and its restrictions do not. As it  can be observed by Definition~\ref{def:abst:df}, decision rules in \SADR\
		have as restriction that an alternative $W$ will belong to $\alternatives{B}$ if conclusion
		$better(Z,W)$ cannot be warranted in \epistcompabs; that is, no alternative
		$Z \in$ \ourB\ exists such that $Z$ \ourRelPrefEst\ $W$.
		The precondition of decision rules belonging to \DRa\ requires that the conclusion $better(W,Y)$ be warranted in \epistcompabs\ ($W$ must be strictly preferred over another alternative $Y \in$~\ourB);
		likewise, the precondition of decision rules belonging to \DRb\ requires that the conclusions $\neg better(W,Y)$ and
		$\neg better(Y,W)$ be warranted in \epistcompabs\ ($W$ must be indifferent wrt. another alternative $Y \in$~\ourB).
		Moreover, by Lemma~\ref{lem:sdast:rat:abs} it can be stated that given the evidence set $E$, if $>_{\mathcal{C}}$ is a strict total order, it will always be possible to warrant the preconditions and/or restrictions of the decision rules in $\Gamma$.
		In this way, in $\alternatives{B}$ there will only be alternatives $x \in$~\ourB\ strictly preferred or indifferent wrt. any other alternative $y \in$~\ourB; that is \mbox{$x$ \ourRelPref~$y$}.
		Hence, $\alternatives{B}=\{ x \in B~|~x \succsim y$ for each $y \in B \}=\MPAB$.
	\end{proof}
\end{theorem}

As mentioned in Section~\ref{sec:theory}, notwithstanding their differences, under certain conditions the PBA and CBA approaches are related.
Below, we present a theorem where the choice behavior of our framework is related with the CBA and its principle of consistency in the decisions made, viz. the weak axiom of revealed preference. To introduce next result, Algorithm \ref{algo:accept} will be considered as the function $\mu(\cdot,\cdot)$ with two arguments: an \textit{abstract decision framework} and a set of alternatives.

\begin{theorem}
	\label{teo:rel:pr:cs:fr:abs}
	Let \ADeF$_\lang$ be an abstract decision framework, where $\epistcompabs=\langle E, W, \bowtie, \mathfrak{pref} \rangle$.
	Given the set \BSet\ which contains all the possible choice experiment, and function $\mu(\cdot,\cdot)$ described in Algorithm~$\ref{algo:accept}$, if $>_{\mathcal{C}}$ is a strict total order  among elements of $\mathcal{C}$, then the choice structure $($\BSet$,\mu(\cdot,\cdot))$ satisfies the weak axiom of revealed preference.
	\begin{proof}
		As stated in Remark~\ref{remark:algo}, we know that function $\mu(\cdot,\cdot)$ described in Algorithm~\ref{algo:accept} implements a choice rule. In this way, it only remains to check that $($\BSet$,\mu(\cdot,\cdot))$ satisfies the restrictions imposed by the weak axiom on the
		choice behavior.
		
		From Lema~\ref{lem:sdast:rat:abs} we know that \epistcompabs\ implements a rational preference relation \ourRelPref, and by Theorem~\ref{teo:pref-rel:abs} it holds that given a choice experiment $B \in \mathcal{B}$, $\alternatives{B}=\MPAB$.
		
		Let us suppose that for some $B \in \mathcal{B}$, it holds that $x,y \in B$ and $x \in \MPAB$; so, from the definition of \MPAB, $x$ \ourRelPref\ $y$.
		To check if the weak axiom holds, let us suppose that for some $B' \in \mathcal{B}$ with $x,y \in B'$ it holds that $y \in \MPAB$.
		This implies that $y$ \ourRelPref\ $z$ for all $z \in B'$; but we already know that $x$ \ourRelPref\ $y$.
		Thus, by transitivity $x$ \ourRelPref\ $z$ for all $z \in B'$, and hence $x \in C^*(B^\prime,\succsim)$.
		This is exactly the conclusion required by the weak axiom of revealed preference.
	\end{proof}
\end{theorem}

As stated in the introductory section, the expected utility theory is a major paradigm in decision making, in spite of its limiting characteristics. The abstract decision framework introduced in previous sections shares some of these characteristics. Both approaches assume: $(i)$ that the decision-maker knows all the options in advance of making the decision; $(ii)$ that options are comparable using whatever criteria the decision-maker uses; $(iii)$ that the decision-maker knows her preferences over these options in advance of making the decision; and $(iv)$ decision rules are independent of the options and preferences, and may be defined in advance.

In this regard, assumptions $(i)$ and $(iii)$ preclude the emergence of options or preferences in the decision making process. In complex domains, deciding preference may be a computationally non-trivial task, and so it is by no means certain that a decision-maker knows her own preferences for all combinations of options.
However, these limitations on our model arise because of the fact that it was conceived in such a way its choice behavior was consistent with the classical approaches described in Section~\ref{sec:theory}.
The main reason we have developed this approach is that we agree with the position stated by Parsons and Fox in~\cite{Par:96}, on the importance of formally relating argumentation-based decision models to classical approaches to decision theory, for the guarantees exhibited in the decisions made. In the following section, where related work is presented, we can see that this formal relation is not accomplished in many recent argumentation-based proposals.

\section{Related work}
\label{sec:related}

The use of argumentative reasoning for decision making has been investigated in other proposals, 
and in this section, we will review those which are more related to our approach.
Most of the proposals to qualitative decision making in argumentation literature (\eg~\cite{Amg:09,Atkinson2006:rev:3,DBLP:conf/ictai/MullerH12,Par:96}) share a common
view with respect to decision making, because they conceive it as a
form of reasoning oriented towards action. That is why, all of them consider
the decision maker's goals or the expected values of actions, to
decide which action to accomplish. This is the main difference with respect to our proposal that is based on the Marketing literature point of view~\cite{RL:91,RL:93}, where each alternative is conceived as a product that the consumer (decision maker) is evaluating to buy (selection).
This approach is detailed next in Subsection~\ref{sec:rw:marketing}.
Then, Subsections~\ref{sec:rw:aabdm} to~\ref{sec:rw:logicsArgumentation} describe main research lines in argumentation-based decision making approaches.
Finally, Subsections~\ref{sec:rw:logicsPreferenceUtility} and~\ref{sec:rw:nonRationalDM} give a more general picture of the decision making problem by connecting our proposal with other relevant approaches, namely, logics of preferences and utilities, and non-rational decision making.

\subsection{Marketing approach to decision making}
\label{sec:rw:marketing}

To the best of our knowledge, the first work on symbolic decision making explicitly following the point of view of the literature on Marketing in decision making was~\cite{collecter2000}, where the application of Defeasible Logic 
for automated negotiation was investigated.
In~\cite{collecter2000}, decision making is performed as a two-stage process formalized in terms of two correlated defeasible theories: the first one for filtering the set of acceptable alternatives based on the buyer's requirements, and the second one, for choosing a particular alternative.
In our proposal, the decision maker is provided with a choice experiment which resembles the set of acceptable alternatives built by the first defeasible theory mentioned above, and we concentrate on choosing what we have called the ``acceptable alternatives'', which would correspond to the chosen ones by the second defeasible theory referred above.

However, the filtering process accomplished by the first defeasible theory of~\cite{collecter2000}, could be naturally modeled with a DAF $\langle
E, W, \bowtie, \mathfrak{pref} \rangle$ as follows: (i) The evidence $E$ would be a consistent set of sentences of the form $a(x)$, such that $x \in X$, $a \in \mathcal{F}$. In this case $X$ refers to the set that is the first component of the ADF introduced in Definition~\ref{def:abst:df}, and $\mathcal{F}$ would be the threshold criteria set that would be used to evaluate alternatives individually. (ii) The working set $W$ will be such that if $x \in X$, $a \in \mathcal{F}$ and $acceptable \not \in \mathcal{F}$ then, for all $w \in W, cl(w)=acceptable(x)~\mathrm{or}~cl(w)=\sim acceptable(x)$ and $pr(w)=\{a(x)\}$. (iii) The conflict relation would remain the same as the one specified in Definition~\ref{def:epist:comp:abs}, and (iv) $\mathfrak{pref}$ relation could be defined analogously as the one of Definition~\ref{def:epist:comp:abs}, by considering an ordering in terms of $\mathcal{F}$ instead of $\mathcal{C}$.

If the proposed decision framework would be used together with the DAF mentioned above to filter alternatives in $X$, the output of this DAF, let us call it $X_f$, should be used instead of $X$ in Definition~\ref{def:abst:df}.
Independently of whether $X$ or $X_f$ is used, a key issue of the decision framework proposed in our work, is that its choice behavior has been formalized with respect to the general theory of choice of Classical Decision Theory and~\cite{collecter2000} has not.

Other works, based on concrete argumentation formalisms, that follow this viewpoint to decision making, are those by Ferretti \etal,~\cite{Fer:08a,JETAI2014}, where Defeasible Logic Proramming (\DLP)~\cite{Gar:04} and Possibilistic Defeasible Logic Programming (P-DeLP)~\cite{Alsi:08}, have been used, respectively. 
These two works could be conceived as particular instances of the abstract decision framework proposed in our work.
The advantage of having an interpreter available like the one built for \DLP~\cite{ksem:07}, is that it makes possible to directly tackle real-world decision-making problems (\eg see \cite{Fer:2010:logos}).
Regarding abstract argumentation frameworks, as far as we know, the one presented here is the first one proposed for decision making having the aforementioned conceptualization of Marketing literature to decision making.

\subsection{Abstract argumentation-based decision making}
\label{sec:rw:aabdm}

A notable abstract argumentation-based framework for decision-making was introduced by Amgoud and Prade in~\cite{Amg:09}, where the decision process within the framework follows two main steps.
First, arguments for beliefs and arguments for options are built and evaluated using classical (Dung's~\cite{Dung:95}) acceptability semantics.
Second, pairs of options are compared using decision principles; these principles are based on the accepted arguments supporting the decisions and they are classified into three categories, whether they consider only arguments in favor or against a decision, both types of arguments, or an aggregation of them into a meta-argument.
This work remains close to the classical view of decision making in that it leaves aside aspects of practical reasoning, such as goal generation, feasibility and planning, to concentrate on the issue of justifying (based on argumentation) the best decision to make in a given situation; besides, it has a logical view of decision that unifies the treatment of multiple criteria decision and decision under uncertainty.
As indicated by Amgoud and Prade, in multiple criteria decision-making each candidate decision $d \in D$ is evaluated from a set $C$ of $m$ different points of view called \textit{criteria}.
Thus, two families of approaches can be distinguished.
On one hand, we have those based on a global aggregation of value criteria-based functions, where the obtained global absolute evaluations are of the form $g(f_1(C_1(d)), \dots , f_m(C_m(d)))$ and the mappings $f_i$ map the original evaluations on a unique scale, which assumes commensurability.
On the other hand, we have the ones that aggregate the preference indices $R_i(d,d')$ into a global preference $R(d,d')$.
Amgoud and Prade follow the former approach while we follow the latter.

An interesting extension to the proposal of Amgoud and Prade, was introduced in~\cite{DBLP:conf/ictai/MullerH12}, where the use of the grounded extension is proposed as acceptability criterion for arguments supporting goals and where a new method for generating decisions is presented.
The proposed approach describes a method for decision analysis in engineering design processes, such as those practiced in the aerospace industry, but also the approach provides support to document the reasons behind decisions for future reference (decision documentation).
A novel issue of this approach is that decisions are modeled as sets of literals rather than as single literal, as usual in argumentation-based decision making literature.
This conceptualization of decisions which may partially overlap, results in a more finely tuned set of decisions when argumentation systems have to be built to derive arguments about the goals achieved when certain decisions are made.
As already mentioned, this proposal extends Amgoud and Prade's approach and hence, differs with our conceptualization of a multi-criteria decision making problem.

\subsection{Assumption-based Argumentation approaches to decision making}

Fan and Toni in~\cite{Fan2014} proposed two different formal frameworks for representing decision making, where this activity is conceived as concerning three related processes: $(a)$ representing information that is relevant to decision making; $(b)$ choosing the decision criteria to represent ``good'' decisions; and $(c)$ computing and explaining the desired decision based on the selected criteria.
This work differ from ours 
in that Fan and Toni use Assumption-based Argumentation (ABA)~\cite{doi:10.1080/19462166.2013.869878}, whereas we use abstract argumentation frameworks; moreover, and more significantly, our approach 
uses pair-wise comparison between decisions to select the ``winning'' decision, whereas in~\cite{Fan2014} a unified process to map decision frameworks into ABA and then compute admissible arguments is developed.
A relation between Fan and Toni's proposal and ours is that both approaches guarantee to choose the best possible decision by constraining the underlying argumentation framework through certain characteristics.
In our case this is attained by defining the epistemic component of the decision framework as a DAF with particular features on the evidence set (\ie the working set), the conflict relation, and preference function (see Definition~\ref{def:epist:comp:abs} for details), and in~\cite{Fan2014} this is accomplished by defining ABA frameworks with different properties so that admissible arguments in those frameworks correspond to strongly dominant, dominant, or weakly dominant decisions in the corresponding mapped decision frameworks, in this way, the decisions made exhibit an element of consistency with respect to the accepted arguments in ABA.
However, as pointed-out by the authors in~\cite{Fan2014} it still remains linking this work to the existing decision theoretic results, which in our case has been accomplished.

Regarding the last observation, Zhong \etal~\cite{DBLP:conf/ecsi/ZhongFTL14}, pushed the proposal in~\cite{Fan2014} one step further to connecting it to existing decision theoretic work; towards that goal, they formally defined a decision ranking mechanism by giving a total order ordering amongst all decisions, but the notion of rationality used relies on the fact that best decisions meet most goals and exhibits fewest redundant attributes (\ie~attributes not contributing to meeting goals).
This leads to the proposal of a new decision criterion, the so-called \emph{minimal deviation}, and its combination with two notions of dominance to select a decision result in the development of two new mappings from decision frameworks onto ABA, in a similar manner as accomplished in~\cite{Fan2014}.
However, the main objective of~\cite{DBLP:conf/ecsi/ZhongFTL14} is the proposal of an algorithm for generating natural language explanations for decisions, on the grounds that existing approaches to argumentation-based decision making either lack automatic support for generating explanations, or directly use the outputs of argumentation engines as explanations.

Another work related to the ABA framework for decision making is that of Matt \etal~\cite{Matt2010}. The differences between this work and~\cite{Fan2014} is that in the latter three different notions of dominant decisions (referred above) have been studied, whereas in the former, only one was studied. 
Besides, one of the frameworks proposed in~\cite{Fan2014} allows the introduction of preferences over goals, while in~\cite{Matt2010} all goals were considered as equally important.
All in all, these works together with other recent papers on ABA frameworks for decision making~\cite{Carstens2015,DBLP:conf/kr/CravenTCHW12,Fan2013}, have shown the suitability of ABA frameworks for dealing with real-world decisions.
In particular, in~\cite{Carstens2015} besides ABA frameworks two other argumentation formalisms are reviewed; namely: Bipolar Argumentation Frameworks~\cite{Cayrol2005} and Value-based Argumentation Frameworks (VAFs)~\cite{Ben:03}.

\subsection{Value-based Argumentation Frameworks for decision making}

Following the work on VAFs~\cite{Ben:03}, in~\cite{Ben:06}, the authors introduced the definition of promoted value, a concept used to define the preference ordering of arguments within the logical formalization developed for Atkinson's analysis of practical syllogism in~\cite{AtkThesis:05}; in this work, the use of argumentation in practical reasoning is studied, proposing a persuasion theory that uses argumentation techniques for obtaining reasons for and against possible actions; then, in~\cite{Atk:06} an application of this approach to a particular medical domain is described.
A significant difference is that, although the approach employs argumentative reasoning to decide among actions,
the work does not focus on a general formalization for decision making through argumentation; also, this approach is integrated in a BDI architecture.

In Bench-Capon \etal~\cite{Bench-Capon2011}, to model agent decision making in experiments in economics, Bench-Capon \etal~proposed a qualitative framework for decision making based on the general argumentation approach to practical reasoning developed in~\cite{Atkinson2006:rev:3}.
This approach is also based on Atkinson's account of practical syllogism~\cite{AtkThesis:05}, where values that are promoted and demoted by alternative actions are explicitly represented by organizing arguments into a VAF.
The main objective of Bench-Capon \etal~is to explore and provide evidence that in economics an argumentative approach of this kind is a better device to explain data from behavioral experiments than the classical approaches to decision making, where agents are expected to be self-interested utility-maximizers.
Towards this goal, the Ultimatum game and the Dictator game were modeled, and then the results were compared with results reported by humans with different cultural background.
The proposed framework is intended for use in situations where agents are required to be adaptable; for instance, where the agent may prefer different outcomes depending on the counter-parties is involved with.
The conceptual difference between this work and our proposal is that we present an argumentation-based decision framework for agents behaving rationally as conceived by classical decision theory, while in~\cite{Bench-Capon2011}, the proposed model is intended to provide a good fit to actual human decision-making processes.

\subsection{The Logic of Argumentation approach to decision making}
\label{sec:rw:logicsArgumentation}

Works such as~\cite{Carstens2015,DBLP:conf/kr/CravenTCHW12,Fan2013,Fan2014,Matt2010,DBLP:conf/ictai/MullerH12} can be considered as good examples of the position stated by Fox \etal~in~\cite{Fox2010831} remarking the importance of developing technologies to support decision making which are grounded on solid theoretical foundations and not only on \emph{ad hoc} methods.
In fact, Fox's position on this matter can be traced to earlier works (see for instance~\cite{Par:96}), where an important conclusion is that when developing decision making models based on argumentation formalisms, a key issue is to formally relate them to classical approaches to decision theory.

In~\cite{Par:96}, Parsons and Fox delineated a proposal in the form of a position paper which advanced the idea that argumentation constitutes a useful framework to reason under uncertainty that can unify different formalisms, such as the possibilistic and probabilistic approaches, and appropriately dealing with inconsistent information; thus, argumentation can become the support of a symbolic model for decision making in practical reasoning.
Later on, Fox and Parsons in~\cite{Fox:97,Fox:98} proposed to use the non-standard logic LA (\underline{L}ogic of \underline{A}rgumentation) developed by Krause~\etal in~\cite{Krause95} in the development of an argumentation system capable of making decisions about the expected values of actions; the resulting approach is similar to the decision theoretic notion of expected value.
The system builds compound arguments, following three stages to construct and combine belief arguments and value arguments.
First, an argument in LA is built supporting that the state associated with a proposition $C$ will occur if action $A$ is taken. Second, a mechanism LV (\underline{L}ogic of  \underline{V}alue) simply assigns a confidence value to $C$.
Finally, a mechanism LEV (\underline{L}ogic of \underline{E}xpected \underline{V}alue) derives arguments over sentences in LA and LV to conclude an expected value for $A$, consistent with the value assigned to $C$.
To choose among alternative actions, the expected value is used to construct a preference ordering over the set of alternative actions.
In this manner, in the context of having sets of arguments that support different actions, the action with highest aggregated value will be selected; this value represents the force of the set of arguments supporting that support that action.

The main idea of our proposal maintains certain similarities to the line of research followed by Fox and Parsons~\cite{Fox:97,Fox:98,Par:96}.
In both approaches, an existing argumentation system for handling belief is chosen as the reasoning formalism to develop a system capable of deciding among competing alternatives.
Nevertheless, our formalization was achieved following different intuitions.
Fox and Parsons built a combined system LA~/~LV~/~LEV where the arguments promote different actions; then, the supporting set of arguments with aggregated value is chosen.
The formalism introduced here relies in an underlying abstract dynamic argumentation framework (DAF) where some components are instantiated and others can be adapted to the application domain, and we have a set of decision rules that work over justified conclusions.

Finally, both approaches are related to Classical Decision Theory but they differ in the manner this relation is accomplished.
Fox and Parsons conceive argumentation as a symbolic model of decision-making and use as underlying argumentation formalism the logic LA, whose theoretic proof method to reason under uncertainty is coherent with
Dempster-Shafer theory.
In our case, the whole design of the framework contributes to get a choice behavior consistent with the general theory of choice of Classical Decision Theory.

\subsection{Logics of preference and utility functions}
\label{sec:rw:logicsPreferenceUtility}

The notion of preference and its role vary in different disciplines; for instance, in the above-mentioned approaches, we find applications of preference principles rather than philosophical foundations.
As mentioned by Hansson~\cite{soh:pl:2001}, the study of general principles for preferences could be traced back to Aristotle's  time, but the first complete systems of preference logics were proposed during the second half of twentieth century~\cite{lob:57,von1972logic}; since then, many contributions have followed this this line of research~\cite{Doyle1991,Doyle94representingpreferences,DBLP:journals/ci/McGeachieD04}. As von Wright~\cite{von1972logic} stated, the existing disagreements on the intuitions about underlying concepts of preference relations of various researchers into this field, can lead to the case where alternative logics of preference can be built in correspondence to these various points of view regarding the matter.

As Doyle and Thomason observed in~\cite{Doyle99backgroundto}, logics of preference could serve as a useful way of organizing and comparing qualitative approaches to decision making.
Our proposal is not the exception to this claim and, in general, many of the concepts and principles used in the framework proposed in the present article are compatible (and could be compared) with similar concepts from logics of preferences.
In the same manner as logics of preference do, our approach allows decision makers more flexibility in expressing incomplete or fewer preferences, or leaving strengths of preferences unspecified~\cite{McGeachieThesis2007}.
However, unlike previous sections where comparisons were focussed on other argumentative approaches, a fair comparison between logics of preference and our proposal would require more general criteria to compare approaches based on very different principles.

McGeachie in~\cite{McGeachieThesis2007} addresses this last issue by proposing an interesting set of criteria that could be used to compare computational properties of (very) different preference representations and their accompanying reasoning methods.
Those criteria include: $(i)$ basic qualitative comparisons, $(ii)$ complex preferences, $(iii)$ utility independence, $(iv)$ (quantitative) tradeoffs, $(v)$ contexts, and $(vi)$ incomplete preferences.
Here, in the context of the logics of preference, a fundamental class of preference representations, termed \emph{ceteris paribus}, is analyzed.
Preference semantics in \emph{ceteris paribus} allow specifying preferences that apply the ``keeping other things equal'' principle, that is, they capture the intuitive idea that unmentioned qualities in preferences might affect the decision making process~\cite{McGeachieThesis2007}.

With respect to the above-mentioned criteria, the \emph{ceteris paribus} assumption proposed in~\cite{Doyle1991} satisfies all of them except the context and tradeoff aspects (criteria $(v)$ and $(vi)$, respectively). In a nutshell, that means that \emph{ceteris paribus} semantics cannot specify that more specific preferences override less specific ones. Besides, they fail in expressing quantitative tradeoffs between variables.
Our proposed framework also exhibits this last limitation in dealing with quantitative tradeoffs; however, context-dependent preferences seem to be naturally captured by our argumentation-based approach.
The direct \emph{ceteris paribus} preferences were later extended with explicit quantification of the ``strength'' of the preference~\cite{TanPearl1994}, graph representations of preferences~\cite{BBCGP1997}, and trade-off ratios between features~\cite{BrafmanD02}.
Most of these extensions dealt with efficiency aspects of  preferences computation or tried alleviating some weaknesses of \emph{ceteris paribus} to fulfill the above-mentioned criteria.

The consideration of \emph{utilities} constitute another mainstream in preference reasoning formalisms which is based on modeling utility functions and computing maximum expected utility, following economic theory of humans as rational decision makers. As pointed out in~\cite{McGeachieThesis2007}, this approach is effective, but can be labor and information-intensive depending heavily on the accuracy of the utility and probability estimates. Traditional economic utility functions succeed in many of the criteria considered above (criteria $(ii)$, $(iii)$, and $(iv)$) but they cannot allow more specific preferences (criterion $(v)$ related to context), are notoriously poor with incomplete information (criterion $(vi)$ incomplete information) and the computation required in many complex situations can be prohibitive~\cite{McGeachieThesis2007}.
In this context, when considering the traditional approaches to Classical Decision Theory, our work mainly differs in that the analysis is directly addressed on the agent's preference relation and not on a utility function that represents this relation,
as usual in these cases.
This is an important feature since to change the preference criteria in our proposal can be easily accomplished by conveniently modifying the order $>_C$.
In contrast, in other approaches using a utility function, this cannot be performed in a direct way or even the whole recalculation of the utility function might be needed; besides, considering the agent's preference relation allowed us to establish a direct connection between our argumentation-based decision-making approach and more essential approaches to modeling individual choice behavior, such as the choice-based approach.

Finally, some recent works have tried to get the best of both, quantitative and qualitative approaches by proposing \emph{hybrid} approaches that, for instance, combine ceteris paribus preferences and utility functions~\cite{McGeachie2005RecompilingUF,DBLP:journals/ci/McGeachieD04,McGeachie2011}.
That is an interesting line of research that we will consider in future works.

\subsection{Non-rational decision making}
\label{sec:rw:nonRationalDM}

The preferences studied in preference logics are usually the preferences of rational individuals but, as stated in~\cite{soh:pl:2001}, they are also used in psychological research where the emphasis is on actual preferences as revealed in behavior. In this respect, at present, there exist several research programs in psychology and behavioral economics based on Simon's criticism of mainstream economic models of perfect rationality (\eg see~\cite{Simon01021955}).
As stated in~\cite{campitelli2010herbert}, the decision making research program in psychology was dominated by Tversky and Kahneman's approach (\eg see~\cite{baron2008thinking,tversky1986rational}) empirically testing Simon's suggestions and showing that they were correct.

Simon's rejection of the assumption of perfect rationality, led him to develop the concept of bounded rationality (\eg see~\cite{Sim:72}).
As suggested in~\cite{Sim:72}, individuals are limited in their rationality for at least these reasons: $(i)$ in order for someone to be rational, she has to fully know and understand the future consequences of her decision-making in the present; $(ii)$ nobody can know in the present the future worth and the impact her actions will have in the future; $(iii)$ in order for someone to be rational she has to know all of the alternatives (usually in decision-making the alternatives someone has in mind are limited and humans are restrained from making optimum decisions).
Considering that our approach to decision making is influenced by the point of view of the literature on Marketing, only point $(iii)$ related to bounded rationality in the choice of alternatives can be related to our work.
In theory, it could be the case that $\mathcal{B}=2^X\setminus\{\emptyset\}$, but in practice the choice experiments presented to the agent will be a significatively smaller set; in this way, in the context of bounded rationality as it was stated by Simon, the rationality of our decision maker is naturally limited to the alternatives the agent has available.\medskip

Finally, it is worth noticing that there is also important research indirectly related to our work. 
For instance, the distinction we make between the epistemic component and the decision component closely resembles Philippe Smets' notions of credal and pignistic probabilities mentioned in~\cite{Smets1994191}, where an interpretation of the Dempster-Shafer model 
is presented as the \emph{transferable belief model}.
Besides, the game-theoretic semantics proposed in~\cite{INT:INT20191} is quite similar to the underlying answer-set semantics of the DAF used to model the epistemic component (see the appendix for details), since both are based on a dialogical view of a process of argumentation that will decide which arguments are finally accepted.
In~\cite{INT:INT20191}, this argumentation process (dialogical deliberation) will be carried out by several agents, while in our case the process will be performed inside the agent's ``mind'' (monological deliberation).

\section{Conclusions and Future Work}
\label{sec:conclusion}

In this work we have introduced an argumentative approach to single-agent decision making.
The proposed formalism, called Abstract Decision Framework, was defined as a tuple  \ADeF\ and involves:
a set $X$ of mutually exclusive alternatives,
a set of distinguished literals \DistLit\
where each $c_i \in \DistLit$  represents a different binary preference relation for comparing alternatives in \SAP,
a strict total order \OrderC\ among elements of \DistLit,
a dynamic argumentation framework  $\epistcompabs$ used for representing preferences relations and conflicts among the available alternatives, and a set of decision rules $\Gamma$ that implements the agent's decision making policy.

Our formalism is not attached to any particular agent architecture, and since is based on abstract argumentation framework, some elements of the formalism can be instantiated with a particular argumentation system.
This is a key issue, given that this abstract argumentation-based framework for decision making can be conceived as a template for building decision frameworks based on concrete argumentation formalisms.
Besides, the resulting concrete instantiation, would result in a decision framework having a choice behavior consistent with the PBA and CBA approaches described in Section~\ref{sec:theory}.

Our proposal takes advantage of the formalism of dynamic argumentation frameworks~\cite{RotsteinMGS10}, where the set of available evidence (that represents the situation of a particular environment where the agent is immersed) activates some arguments of the framework, and those arguments are used for computing the agent's justified conclusions with respect to that evidence.
For deciding between conflicting arguments, the formalization relies in an order over the preference criteria defined for the application domain.

In this article, we focused on the ``argumentation-based view'' to decision making and establishing formal connections to Classical Decision Theory.
However, as we saw in Section~\ref{sec:rw:logicsPreferenceUtility}, other qualitative, quantitative, and hybrid approaches have played a key role in decision making research.
As future work, we plan to establish explicit connections with those influential approaches; our studies will include a formal comparison with other qualitative representations such as multi-attribute \emph{ceteris paribus} preference statements~\cite{Doyle1991,Doyle94representingpreferences} and semantics for extensions that support quantitative comparisons involving tradeoffs~\cite{DBLP:journals/ci/McGeachieD04,McGeachie2005RecompilingUF,McGeachie2011}.
In this context, an important aspect to analyze is the computational burden involved in recompiling utility functions in some recent hybrid approaches~\cite{McGeachie2005RecompilingUF} versus the flexibility provided by argumentation approaches that support numeric strengths in their arguments in the context of changing environments.
Therefore, to get some insights on this matter, it would be interesting to model the decision making formalism presented in~\cite{JETAI2014} (based on P-DeLP) as an instance of our abstract framework.
Also, to test its adequacy as an abstract decision framework, other instantiations will be performed with argumentative formalisms having semantics resembling a dialogical discussion; for instance~\cite{Prak:97}.

Moreover, as aforementioned in Section~\ref{sec:rw:nonRationalDM}, new research lines arising from psychology and behavioral economics are taking into account non-rational behavior in decision making based on Simon's seminal works (\eg see~\cite{Simon01021955,Sim:72}).
Nevertheless, as stated in~\cite{campitelli2010herbert}, perfect rational models are still the prevailing models; hereof, we have formalized the choice behavior of the proposed framework with respect to the general theory of choice of Classical Decision Theory (as motivated by Parsons and Fox in~\cite{Par:96} and already mentioned in Sections~\ref{sec:intro}~and~\ref{sec:related}).
As future work, following the thorough presentation in~\cite{campitelli2010herbert}, we will also study how our framework can be redefined to model some of the proposals concerning non-rational decision making, given that argumentation-based approaches are alike to human reasoning.

\section*{Acknowledgments}
This work was partially supported by PGI-UNS (grants 24/N035, 24/N030), PIP-CONICET (grant 112-201101-01000), Universidad Nacional de San Luis (PROICO 30312)
and EU H2020 research and innovation programme under the Marie Sklodowska-Curie grant agreement No. 690974 for the project MIREL: MIning and REasoning with Legal texts.

\section{Appendix: Dynamic Argumentation Framework (DAF)}
\label{sec:abst:nico}

We will introduce here the main concepts of the \emph{Dynamic Argumentation Framework} (DAF) proposed by Rotstein~\etal in~\cite{RotsteinMGS08,RotsteinMGS10}; this framework was used in Section~\ref{sec:framework} to  formalize our abstract decision framework.
DAFs have been  built as a refinement of Dung's argumentation framework (AF)~\cite{Dung:95}, which is defined as a pair containing a set of arguments and a defeat relation ranging over pairs of them.
The objective of the approach proposed in~\cite{RotsteinMGS08,RotsteinMGS10} was to extend Dung's AFs to handle dynamics.
To cope with this, it was considered a set of available evidence, which determines which arguments can be used to make inferences.
In Dung's approach, the consideration of a changing set of arguments would involve passing from one framework to another.

In DAFs, the notion of \emph{evidence} is a key concept.
Evidence is considered to be an entity represented on a logical language, denoted as $\mathcal{L}$, whose sentences correspond to facts in the domain, also this language will be used as the base language to represent the arguments' \textit{premises} and \textit{conclusions}.
The arguments supported by the current evidence will be considered as active; on the other hand, the set of evidence contains those sentences that are undisputed in the situation at hand.
Likewise, arguments are considered the smallest reasoning steps to be represented in the DAF; that is, the smallest piece of reasoning that provides backing for a claim from a set of premises, as formally stated in Definition~\ref{def:argu:nico}.
The complement notation will be used to state that a sentence is the negation of another one:
$\overline{\alpha}=\neg \alpha,~\overline{\neg \alpha}=\alpha$.

\begin{definition}
	\label{def:argu:nico}
	Given a language $\mathcal{L}$, an \emph{argument} $\mathcal{A}$ is a reasoning step for
	a claim $\alpha \in \mathcal{L}$ from a non-empty set of premises $\{\beta_1, \ldots, \beta_n\} \subseteq \mathcal{L}$ such that $\beta_i \neq \alpha, \beta_i \neq \overline{\alpha},$ and $\beta_i \neq \overline{\beta_j}$ for every $i, j, 1 \leq i, j \leq n$.
\end{definition}

Given an argument $\mathcal{A}$, we will identify both its claim and its set of premises through the functions \emph{cl}($\mathcal{A}$) and \emph{pr}($\mathcal{A}$), respectively. Given \emph{pr}($\mathcal{A}$) = $\{\beta_1, \ldots, \beta_n\}$ and \emph{cl}($\mathcal{A}$) = $\alpha$, the {\em interface} of $\mathcal{A}$ is denoted as $\inti{\mathcal{A}}=\langle \{\beta_1, \ldots, \beta_n\}, \alpha \rangle$.
In a DAF, given an evidence set $E$, an argument's premise is satisfied if it belongs to $E$, or it is the conclusion of an active argument according to $E$; an argument will be considered \emph{active} if its premises are satisfied.
Notice that an argument $\mathcal{A}$ not having enough evidencial support for all its premises may have active supporting arguments for the remaining unsatisfied premises, \ie~active arguments whose claims are the unsatisfied premises of $\mathcal{A}$.
An argument $\mathcal{B}$ is a \emph{supporting argument} of an argument $\mathcal{A}$ iff $cl(\mathcal{B}) \in pr(\mathcal{A})$.
Let $cl(\mathcal{B})=\beta$, then we say that $\mathcal{B}$ {\em supports} $\mathcal{A}$ through $\beta$.

The DAF imposes as a requirement that all the arguments be coherent.
The concept of a coherent argument is related to the fact that certain restrictions must be addressed in order to consider an argument as a consistent reasoning step.

\begin{definition}
	\label{def:argu:coher:nico} An argument $\mathcal{A}$ is {\bf coherent} wrt. a set $E$ of evidence iff $\mathcal{A}$ verifies:
	\begin{itemize} \itemsep -1pt
		\item Internal Consistency: $\overline{cl(\mathcal{A})} \notin pr(\mathcal{A})$;
		\item Consistency wrt. $E$: $\overline{cl(\mathcal{A})} \notin E$;
		\item No Internal Redundancy: $cl(\mathcal{A}) \notin pr(\mathcal{A})$;
		\item No Redundancy wrt. $E$: $cl(\mathcal{A}) \notin E$.
	\end{itemize}
\end{definition}

\begin{definition}
	\label{def:argu:activ:nico} Given a set $Args$ of arguments and a set $E$ of evidence, an argument $\mathcal{A}
	\in Args$ is {\bf active} wrt. $E$ iff $\mathcal{A}$ is coherent and for each $\beta \in$~pr($\mathcal{A}$)
	either $\beta \in E$, or there is an active argument $\mathcal{B} \in Args$ that supports $\mathcal{A}$ through $\beta$.
\end{definition}

\begin{definition}
	Given a set $Args$ of arguments, an {\bf argumental
		structure} for a claim $\alpha$ from $Args$ is a tree of arguments $\Sigma$ verifying:
	\begin{enumerate}
		\item The root argument $\mathcal{A}_t \in Args$, called top argument, is such that $cl(\mathcal{A}_t) = \alpha$
		and is noted as $top(\Sigma)$;
		\item A node is an argument $\mathcal{A}_i \in Args$ such that for each premise $\beta \in pr(\mathcal{A}_i)$ there
		is at most one child argument in $Args$ supporting $\mathcal{A}_i$ through $\beta$;
		\item A leaf is an argument $\mathcal{A}_h \in Args$ such that there is no argument in $Args$ supporting it.
	\end{enumerate}
\end{definition}

Regarding notation for an argumental structure $\Sigma$:\begin{itemize}
	\item The set of arguments belonging to $\Sigma$ is denoted as $args(\Sigma)$.
	\item The set of premises of $\Sigma$ is: $pr(\Sigma) =
	\stackbin[\mathcal{A}~\in~args(\Sigma)]{}{\bigcup}pr(\mathcal{A})
	\setminus
	\stackbin[\mathcal{A}~\in~args(\Sigma)]{}{\bigcup}cl(\mathcal{A})$.
	\item The claim of $\Sigma$ is denoted as $cl(\Sigma) = \alpha$.
\end{itemize}

Functions $\mathit{pr}(\cdot)$ and $\mathit{cl}(\cdot)$ are also applied to argumental structures.
It is important to stress that, within an argumental structure, if a premise which is unsupported by the evidencial set appears more than once, that premise must be supported by the same argument.
Moreover, when the set of arguments in an argumental structure is a singleton, the argumental structure is called \emph{primitive}.
In DAF, the defeat relationship is obtained by applying a preference function  on pairs of argumental structures in conflict.
As mentioned above, arguments are the finest grained steps in the reasoning process, so it makes sense to define the conflict relationship on arguments first, and then to extend it to argumental structures.
This gives rise to the notion of conflict among argumental structures; when all these  conflicts are determined, a preference function decides which  argument (supported by an argumental structure) prevails.
Given a set of arguments $Args$, the set $\bowtie \subseteq Args \times Args$ denotes a
{\em conflict relation} over $Args$, verifying $\bowtie\supseteq\{(\mathcal{A},\mathcal{B})|\{\mathcal{A},\mathcal{B}\}\subseteq
Args, cl(\mathcal{A})=\overline{cl(\mathcal{B})}\}$.
Given a set $Args$ of arguments, an argument $\mathcal{A}_i$
{\em transitively supports} an argument $\mathcal{A}_k$ within $Args$ iff there is a sequence of arguments $[\mathcal{A}_i,\ldots,\mathcal{A}_k]$ in $Args$ where $cl(\mathcal{A}_j) \in
pr(\mathcal{A}_{j+1})$, for every $j$ such that $i \leq j \leq k-1$.
\begin{definition}
	\label{def:EA:wf}
	Let $\mathit{Args}$ be a set of arguments and $\bowtie\,\subseteq \mathit{Args} \times \mathit{Args}$ be a conflict relation, an argumental structure $\Sigma \in \mathit{Args}$ is {\em well-formed} wrt. $\bowtie$, iff $\Sigma$ verifies:
	\begin{itemize}
		\item Premise Consistency: There are no $\alpha, \beta \in \textrm{pr}(\Sigma)$ such that $\overline{\alpha} = \beta$;
		\item Consistency: For each argument $\mathcal{A} \in \mathit{args}(\Sigma)$ there is no argument $\mathcal{B} \in
		\mathit{args}(\Sigma)$, such that $\mathcal{A} \bowtie \mathcal{B}$.
		\item Non-Circularity: No argument $\mathcal{A} \in \mathit{args}(\Sigma)$ transitively supports an argument $\mathcal{B} \in \mathit{args}(\Sigma)$ if $\mathit{cl}(\mathcal{B}) \in \mathit{pr}(\mathcal{A})$.
		\item Uniformity: If $\mathcal{A} \in \mathit{args}(\Sigma)$ supports  $\mathcal{B} \in \mathit{args}(\Sigma)$
		through $\beta$, then for all $\mathcal{B}_i \in
		\mathit{args}(\Sigma)$ having $\beta$ as premise, $\mathcal{A}$
		supports $\mathcal{B}_i$ through $\beta$.
	\end{itemize}
\end{definition}

The domain of all well-formed argumental structures wrt. $\mathit{Args}$ and $\bowtie$ is denoted as $\mathfrak{str}_{(\mathit{Args},\bowtie)}$.
Since a set of evidence is always consistent, a structure with inconsistent premises would never become active.
However, as stated above, it is useful to validate also inactive argumental structures.
The property of consistency invalidates inherently contradictory argumental structures.
The requirement of non-circularity avoids taking into consideration structures yielding infinite reasoning chains.
Finally, the restriction of uniformity does not allow heterogeneous support for a premise throughout a structure.
These constraints are defined so that we can trust a well-formed structure as a sensible reasoning chain,
independently from the set of evidence.

From a knowledge representation perspective, the objective of an argumental structure is comparable to that of its composing arguments: both support claims from a set of premises. The difference relies on the fact that arguments cannot be decomposed into smaller pieces. Conversely, argumental structures can be decomposed into subsets (aggregations of arguments) referred as argumental substructures, as defined below.
Given two argumental structures $\Sigma$ and $\Sigma'$ from a set of arguments $\mathit{Args}$, $\Sigma'$ is an
{\em argumental substructure} of $\Sigma$ (denoted as $\Sigma' \sqsubseteq \Sigma$) iff $args(\Sigma') \subseteq
args(\Sigma)$.
If $args(\Sigma') \subset args(\Sigma)$, then $\Sigma'$ is a proper argumental substructure of $\Sigma$
(denoted as $\Sigma' \sqsubset \Sigma$).
Given a set $\mathit{Args}$ of arguments, a conflict relation $\bowtie \subseteq \mathit{Args} \times \mathit{Args}$, and two argumental structures $\Sigma_1,\Sigma_2 \in \mathfrak{str}_{(\mathit{Args},\bowtie)}$, structure $\Sigma_1$ is in {\em conflict} with $\Sigma_2$ iff $top(\Sigma_1) \bowtie top(\Sigma_2)$.
The conflict between argumental structures is denoted as ``\conflictEA''.

The preference function is defined over argumental structures and not over arguments, since, in order to decide which argument prevails, all the knowledge giving support to them should be considered.
Moreover, when facing different scenarios, the same argument might be active from different active argumental structures and, consequently, the preference could change along with evidence.

Given two argumental structures $\Sigma_1,\Sigma_2 \in
\mathfrak{str}_{(\mathit{Args},\bowtie)}$, the {\em preference function}
\mbox{$\mathfrak{pref} : \mathfrak{str}_{(Args,\bowtie)} \times
	\mathfrak{str}_{(\mathit{Args},\bowtie)} \mapsto
	\mathfrak{str}_{(\mathit{Args},\bowtie)} \cup \{\epsilon\}$} on argumental structures,
\mbox{$\mathfrak{pref}(\Sigma_1,\Sigma_2) = [\Sigma_1|\Sigma_2|\epsilon]$},
determines the preferred argumental structure; if none is preferred, the function returns the constant $\epsilon$.

\begin{definition}
	\label{def:EA:derrota}
	Given two argumental structures $\Sigma_1,\Sigma_2  \in \mathfrak{str}_{(\mathit{Args},\bowtie)}$, $\Sigma_1$ {\em defeats} $\Sigma_2$, iff there is an argumental substructure $\Sigma \sqsubseteq \Sigma_2$, such that $\Sigma_1$~\conflictEA~$\Sigma$ and $\mathfrak{pref}(\Sigma_1,\Sigma)=\Sigma_1$.
	The defeat relation between argumental structures is denoted as ``$\,\Rightarrow$''.
\end{definition}

When a structure defeats another, the attack comes from the claim of the former to any claim of a substructure of the latter. The attack is not directed to an argument, but to
a substructure, which is the actual portion of the structure under attack.

\begin{definition}
	\label{def:daf:def:nico} A {\em dynamic argumentation framework (DAF)} is a four tuple $\langle E, W, \bowtie,
	\mathfrak{pref}\rangle$, composed by a set $E$ of evidence, a working set $W$ of arguments, a conflict relation $\bowtie\,\subseteq W \times W$, and a preference function $\mathfrak{pref}$ defined over $\mathfrak{str}_{(W,\,\bowtie)}$.
\end{definition}

The working set of arguments contains every argument that is available for use by the reasoning process.
At any given moment, the set of active arguments will represent the current situation.
Different instances of the set of evidence will determine different instances of the DAF; thus, when ``restricting'' a DAF to its associated set of evidence, we obtain an abstract argumentation framework AF in the classical sense~\cite{Dung:95}, \ie~a duple with the set of active arguments, and the attack relation between pairs of active arguments.
This ``restriction'' is called an active instance and it is addressed in Definition~\ref{def:daf:ins:ac}. 
Below, other necessary definitions are introduced first.
\begin{definition}
	\label{def:conj:argu:activ} Given a DAF $\mathbf{F} = \langle E, W, \bowtie, \mathfrak{pref}\rangle$, the {\em set of active argumental structures} in $\mathbf{F}$ wrt. $E$ is
	$\mathbb{S} = \{\mathcal{A} \in W~|~\mathcal{A}~\mathrm{is~active~wrt.}~E\}$.
\end{definition}
\begin{definition}
	\label{def:EA:activa} Given a DAF $\mathbf{F} = \langle E, W, \bowtie, \mathfrak{pref}\rangle$, a {\em well-formed  argumental structure} $\Sigma$ in $\mathbf{F}$ is {\em active} wrt. $E$ iff $pr(\Sigma) \subseteq E$ and each argument in
	$args(\Sigma)$ is coherent wrt. $E$.
\end{definition}
\begin{definition}
	Given a DAF $\mathbf{F} = \langle E, W, \bowtie, \mathfrak{pref}\rangle$ and the set $\mathbb{S}$ of active argumental structures in $\mathbf{F}$ wrt. $E$, the {\em active defeat relation} over argumental structures in $\mathbf{F}$ is \mbox{$\derrEAact = \{(\Sigma_1,\Sigma_2) \in\ \Rightarrow\,|~\Sigma_1,\Sigma_2 \in \mathbb{S}\}$}.
\end{definition}
\begin{definition}
	\label{def:daf:ins:ac} Given a DAF $\mathbf{F} = \langle E, W, \bowtie, \mathfrak{pref}\rangle$, the {\em active instance} of $\mathbf{F}$ is the abstract argumentation framework AF $(\mathbb{S},\derrEAact)$, where $\mathbb{S}$ is the set of the active argumental structures from $W$ wrt. $E$, and
	\derrEAact\ is the active attack relation between structures in $\mathbb{S}$.
\end{definition}

Every DAF, at any given moment, has an associated active instance, \ie~an abstract argumentation framework AF; therefore, all the work done on acceptability of arguments and argumentation semantics for the abstract frameworks can be applied to the DAF just by finding from its active instance the set of accepted argumental structures.
Moreover, since structures hold a claim, we can go a step further and consider justification of claims, either skeptically or cautiously.
In this way, the study of semantics on the DAF relating arguments, structures and substructures, it is not only faced calculating the active instance but tools are also defined to directly calculate semantics in the DAF.
With this aim, as detailed next, dialectical trees are used.

A natural choice to obtain the supported conclusions from a set of arguments is to resort to one of the many possible semantics developed for abstract argumentation frameworks.
In his landmark work, Dung~\cite{Dung:95} described \emph{preferred, stable, grounded} and \emph{complete} semantics based on the notion of the admissibility; subsequently, other semantics where introduced: \emph{ideal, semi-stable, stage,} and \emph{CF}2, although the last two are not based on admissibility (see~\cite{Baroni2011} for an excellent introduction to the topic of argumentation semantics).
An interesting observation is that our formalism requires a skeptical posture regarding the inferential result; that is to say, if the semantics chosen is not skeptical, 
we would need to consider those arguments that appear in all the extensions of the framework under scrutiny.
In this paper, as an alternative semantics, we have introduced an answer-set semantics which is based on a dialogical view of the process of argumentation that will decide which arguments are finally accepted; this approach will be formulated below.

An argumentative approach based on dialectical trees corresponds to the intuition of a discussion around a topic.
That is, the argumental structure on the top of the tree is not only supporting a claim $\alpha$, but also a topic of discussion. In this respect, it will be discussed in terms of $\alpha$, $\neg \alpha$ and additionally, the topics introduced by all the argumental structures involved in supporting sentences, since they contain substructures that might be attacked (see Definition~\ref{def:EA:derrota}).
Next, all the basic notions related to argument interactions are described to address the formal definition of dialectical tree.

Given a DAF $\mathbf{F}$, an {\em argumentation line} is a sequence $\lambda=[\Sigma_1,\ldots,\Sigma_n]$, where each $\Sigma_i~(1 \leq i \leq n)$ is a well-formed argumental structure in $\mathbf{F}$ attacking its predecessor.
The \emph{root} of $\lambda$ is $\Sigma_1$ and the \emph{leaf} is $\Sigma_n$.
Given an argumentation line $\lambda=[\Sigma_1,\ldots,\Sigma_n]$, the {\em top segment} of
\mbox{$\Sigma_i~(1 < i \leq n)$} in $\lambda$ is $[\Sigma_1,\ldots,\Sigma_i]$ and it is denoted as
$\lambda^{\uparrow}(\Sigma_i)$.
The {\em proper top segment} of $\Sigma_i$ in $\lambda$ is $[\Sigma_1,\ldots,\Sigma_{i-1}]$ and is denoted as $\lambda^{\uparrow}[\Sigma_i]$.

As already mentioned, the exchange of arguments resembles a dialogical discussion; as such, it makes sense that the introduction of a new argument by one of the participants should be consistent with her previously posed arguments.
Indeed, it is also desirable to require that none of the parties be allowed to introduce an argument already posed by them.
We will refer as \emph{pro} and \emph{con} each of the parties involve in the dialogical argumentation process.
The following definition precedes the formalization of the intuitions referred above on an \emph{acceptable argumentation line}.
Given a set $S$ of structures and a conflict relation \conflictEA~$\subseteq S \times S$, the {\em set} $S$ is {\em consistent} wrt. \conflictEA, iff there is no $\{\Sigma_1,\Sigma_2\}\subseteq S$ such that $\Sigma_1$~\conflictEA~$\Sigma_2$.
Given an argumentation line
$\lambda=[\Sigma_1,\ldots,\Sigma_n]$, the {\em set pro (con) of argumental structures} is composed by all the $\Sigma_i~(1 \leq i \leq n)$, with odd (even) $i$ values.
The set pro (con) of structures in $\lambda$ is denoted $\lambda^+$ ($\lambda^-$).
Given an argumentation line $\lambda$ within the context of a DAF $\mathbf{F}=\langle E, W,
\bowtie, \mathfrak{pref}\rangle$, $\lambda$ is {\em acceptable} in $\mathbf{F}$ iff the following restrictions hold:
(1)~Non circularity: there is no repetition of structures in $\lambda$, and
(2)~Concordance: sets \emph{pro} and \emph{con} are consistent wrt. \conflictEA.
It is worth mentioning that an acceptable argumentation line is \emph{exhaustive} if it is not possible to insert more
argumental structures in the sequence.

\begin{definition}
	\label{def:arb:dial:nico} Given a DAF $\mathbf{F}$ and a set  $S$ of exhaustive argumentation lines in $\mathbf{F}$ rooted in $\Sigma_1$, such
	that $S$ is maximal wrt. set inclusion, a
	{\em dialectical tree} for an argumental structure $\Sigma_1$
	is a tree $\mathcal{T}_{\mathbf{F}}(\Sigma_1)$
	verifying:
	\begin{itemize}
		\item $\Sigma_1$ is the \emph{root};
		\item A structure $\Sigma_{i \neq 1}$ in a line $\lambda_i \in
		S$ is an \emph{inner node}, iff has as children all the  $\Sigma_j$ in lines $\lambda_j \in S$ such that $\Sigma_j
		\Rightarrow \Sigma_i$ and
		$\lambda^{\uparrow}[\Sigma_i]=\lambda^{\uparrow}(\Sigma_j)$;
		\item The \emph{leaves} of the tree correspond to the leaves of the lines in $S$.
	\end{itemize}
\end{definition}

Dialectical trees are defined over the working set of arguments, and hence they can contain active and inactive
argumental structures. A dialectical tree that contains only
active structures is called \emph{active dialectical tree}, and it is denoted $\mathbb{T}_{\mathbf{F}}(\Sigma)$.
Once a dialectical tree has been built for an argumental structure, a marking criterion determines which structures in the tree are defeated and which ones remain undefeated.
This criterion is specified by a marking function. In~\cite{Rot:09} different marking approaches are described. As stated next, a structure can be marked as ``defeated'' or ``undefeated''.
Given a DAF $\mathbf{F}$ and an argumental structure $\Sigma_i$ in a line $\lambda_i$ in a
dialectical tree $\mathcal{T}_{\mathbf{F}}(\Sigma)$, a {\em marking function} $\mathfrak{m}$ is
$\mathfrak{m}(\Sigma_i,\lambda_i,\mathcal{T}_{\mathbf{F}}(\Sigma))=[U|D]$,
where $U$ represents ``undefeated'' and $D$ ``defeated''.
A choice for this marking function is the one defined in \DLP~\cite{Gar:04}, where the evaluation is done in a skeptical manner; that is, an argumental structure is considered as undefeated only when all its defeaters have been defeated.
Given a dialectical tree
$\mathcal{T}_{\mathbf{F}}(\Sigma)$, the {\em skeptical marking function} $\mathfrak{m}_e$ is defined as follows:
\noindent $\mathfrak{m}_e(\Sigma_i,\lambda_i,
\mathcal{T}_{\mathbf{F}}(\Sigma)) = D$ \emph{iff} $\exists
\Sigma_j~\mathfrak{m}_e(\Sigma_j,\lambda_j,
\mathcal{T}_{\mathbf{F}}(\Sigma)) = U$, where $\Sigma_j$ is a child of $\Sigma_i$ in $\mathcal{T}_{\mathbf{F}}(\Sigma)$.
Once the marking function has been defined, the warranty status of the root of a dialectical tree can be determined, as defined next.

\begin{definition}
	\label{def:warrant:nico} Given a DAF $\mathbf{F}$ and a marking function $\mathfrak{m}$, an argumental structure $\Sigma$ from
	$\mathbf{F}$ is {\em warranted} in $\mathbf{F}$, iff
	$\mathfrak{m}(\Sigma,\lambda,\mathbb{T}_{\mathbf{F}}(\Sigma)) = U$,
	where $\lambda$ is any argumentation line from
	$\mathbb{T}_{\mathbf{F}}(\Sigma)$. The conclusion $cl(\Sigma)$ is \emph{justified} by $\mathbf{F}$.
\end{definition}

Naturally, if a different marking function is considered, the
definition of warrant consequently changes.
The warranted argumental structures will be determined by
Definition~\ref{def:warrant:nico}, giving rise to the notion of semantics on the DAF.
It is worth mentioning that the notion of warrant is defined on active dialectical trees, since all the reasoning only can  be carried out over the set of active arguments.


\bibliography{aij}
\bibliographystyle{plain}

\end{document}